\documentclass[11pt]{article}
\usepackage[final]{acl}
\usepackage{times}
\usepackage{latexsym}
\usepackage[T1]{fontenc}
\usepackage[utf8]{inputenc}
\usepackage{microtype}
\usepackage{inconsolata}
\usepackage{graphicx}
\usepackage{amsmath}
\usepackage{subcaption}
\usepackage{float}
\usepackage{booktabs}
\usepackage{multirow}
\usepackage{lineno}
\usepackage{tcolorbox}
\usepackage{pifont}
\usepackage{xcolor}
\usepackage{makecell}
\usepackage{tikz}
\usepackage{caption}
\usepackage{enumitem}
\usepackage{amssymb}
\usepackage{algorithm}
\usepackage{algpseudocode}
\usepackage{authblk}
\title{Heterogeneous Adaptive Policy Optimization: Tailoring Optimization to Every Token's Nature}

\makeatletter
\renewcommand\AB@affilsepx{, \protect\Affilfont}
\makeatother

\author[1,2,*]{\textbf{Zheng~Liu}}
\author[1,2,*]{\textbf{Mengjie~Liu}}
\author[3]{\textbf{Siwei~Wen}}
\author[2]{\textbf{Mengzhang~Cai}$^2$, \textbf{Bin~Cui}$^{1,5}$, \\ \textbf{Conghui~He}}
\author[1,4,5,†]{\textbf{Wentao~Zhang}}
\affil[1]{Peking University} \affil[2]{Shanghai AI Laboratory} \affil[3]{Beihang University} \affil[4]{Zhongguancun Academy, \protect\\ \textsuperscript{5}Beijing Key Laboratory of Software and Hardware Cooperative Artificial Intelligence Systems}

\begin{document}
\maketitle

\let\thefootnote\relax
\footnotetext{* Equal contribution.}
\footnotetext{† Corresponding authors.}

\begin{abstract}
Using entropy as a measure of heterogeneity to guide optimization has emerged as a crucial research direction in Reinforcement Learning for LLMs. However, existing methods typically treat it as a discrete filter or post-hoc regulator rather than a core optimization driver. 
To fully leverage the potential of entropy and achieve fine-grained regulation, we introduce \textbf{H}eterogeneous \textbf{A}daptive \textbf{P}olicy \textbf{O}ptimization (HAPO), a token-aware algorithm that continuously adapts optimization dynamics based on token-level entropy throughout the entire training process. Our algorithm includes four key components: (1) \textbf{Adaptive Temperature Sampling} that adjusts sampling temperature in real time, promoting exploration at high-entropy tokens. (2) \textbf{Token-Level Group Average Advantage Estimation} that estimates advantages at token level, accounting for sequence-length effects while preserving non-biased treatment.(3) \textbf{Differential Advantage Redistribution} that leverages entropy and importance ratios to adjust advantages for tokens with clear signals. (4) \textbf{Asymmetric Adaptive Clipping} that adynamically adjusts clipping boundaries based on token-level entropy. Through systematic investigation of entropy, we embed token-level treatment into every stage. Extensive experiments on mathematical reasoning, code, and logic tasks across multiple models demonstrate HAPO's consistent superiority over DAPO. Our code can be found in \href{https://github.com/starriver030515/HAPO}{https://github.com/starriver030515/HAPO}.

\end{abstract}
\section{Introduction}
\label{sec:intro}
Reinforcement Learning from Human Feedback~\citep{ouyang2022traininglanguagemodelsfollow} has emerged as a fundamental technique for enhancing the reasoning capabilities of LLMs. State-of-the-art models including OpenAI o1~\citep{o1}, DeepSeek-R1 ~\citep{deepseekai2025deepseekr1incentivizingreasoningcapability}, and Qwen3 series ~\citep{yang2025qwen3technicalreport} have achieved remarkable performance gains, underscoring how carefully designed reinforcement learning frameworks can empower LLMs to tackle tasks requiring reasoning.

\begin{figure*}[!t]
    \centering
    \includegraphics[width=15cm]
    {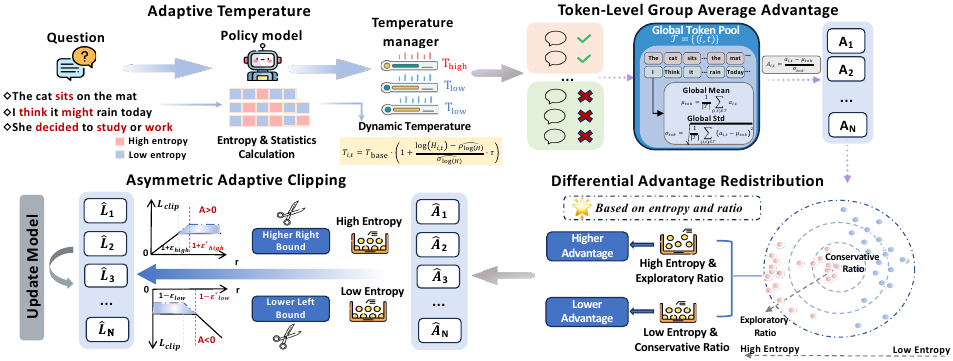}
    \caption{HAPO's overall framework. HAPO integrates Adaptive Temperature Sampling, Token-Level Group Average Advantage Estimation, Differential Advantage Redistribution and Asymmetric Adaptive Clipping, leveraging entropy as a core optimization driver to achieve fine-grained, token-level optimization.}
    \label{fig:overall_arch}
\end{figure*}

However, existing algorithms~\citep{schulman2017proximalpolicyoptimizationalgorithms,shao2024deepseekmathpushinglimitsmathematical,yu2025dapoopensourcellmreinforcement,fang2026proximity,fu2026maspo} share a fundamental limitation: they employ a uniform optimization across all tokens, failing to distinguish between tokens that represent critical reasoning paths vs routine patterns. This conflicts with the heterogeneous nature of language generation, where tokens serve different roles in the reasoning.

Recent works have pioneered token heterogeneity by using entropy to distinguish tokens with different roles.  DAPO with forking tokens ~\citep{wang20258020rulehighentropyminority} proves that only a minority of high-entropy tokens guide the optimization process. Archer ~\citep{wang2025stabilizingknowledgepromotingreasoning} divides tokens into high/low entropy groups, relaxing clipping bounds for high-entropy tokens to encourage exploration. However, these methods function entropy as discrete filters, failing to leverage it for fundamental optimization and adjustment. Other methods leverage entropy to adjust advantages. Entropy Adv~\citep{cheng2025reasoningexplorationentropyperspective} adds an entropy term in advantage to encourage extended reasoning, while EDGE-GRPO~\citep{zhang2025edgegrpoentropydrivengrpoguided} utilizes sequence-level entropy to reweight advantages in favor of confident outputs. While realizing a form of heterogeneity, they restrict entropy to an auxiliary regulator or post-hoc bonus, leaving the core optimization mechanics fundamentally unchanged.

These limitations motivate us to develop an algorithm that embeds entropy into the core of optimization, thereby fully leveraging its potential and achieve fine-grained regulation. We first conduct a systematic empirical analysis, delving into the mechanisms and requirements for heterogeneous treatment across different training stages:

\textbf{\textit{Rollout Generation: Exploration-Exploitation Imbalance.}} High-entropy
tokens encoding critical reasoning decisions are rare. We find the temperature dilemma: low temperatures preserve accuracy but suppress critical tokens, while high temperatures increase their occurrence but introduce noise.

\textbf{\textit{Advantage Calculation: Reward Attribution Granularity.}} Standard sequence-level advantage estimation ignores intra-sequence token heterogeneity and becomes problematic under DAPO's token-mean averaging loss where longer negative samples dominate gradients. Moreover, our importance ratio analysis shows that tokens require different update magnitudes regardless of their entropy values, yet uniform and entropy-based reweighting fail to identify critical reasoning decisions.

\textbf{\textit{Clipping Loss Computation: Clipping Constraint Mismatch.}} Our clipping analysis reveals paradoxical patterns: low-entropy tokens (mostly formatting symbols) hit left bounds preventing probability reduction, while high-entropy tokens (crucial reasoning elements) hit right bounds limiting exploration. Uniform clipping thus protects noise while constraining exploration.

Based on the discoveries above, we introduce \textbf{Heterogeneous Adaptive Policy Optimization (HAPO)}. Moving beyond simple regulation, HAPO formulates optimization parameters as continuous functions of entropy to embed fine-grained treatment into every stage:

\begin{figure*}[!t]
    \centering
    \includegraphics[width=16cm,height=3.6cm]{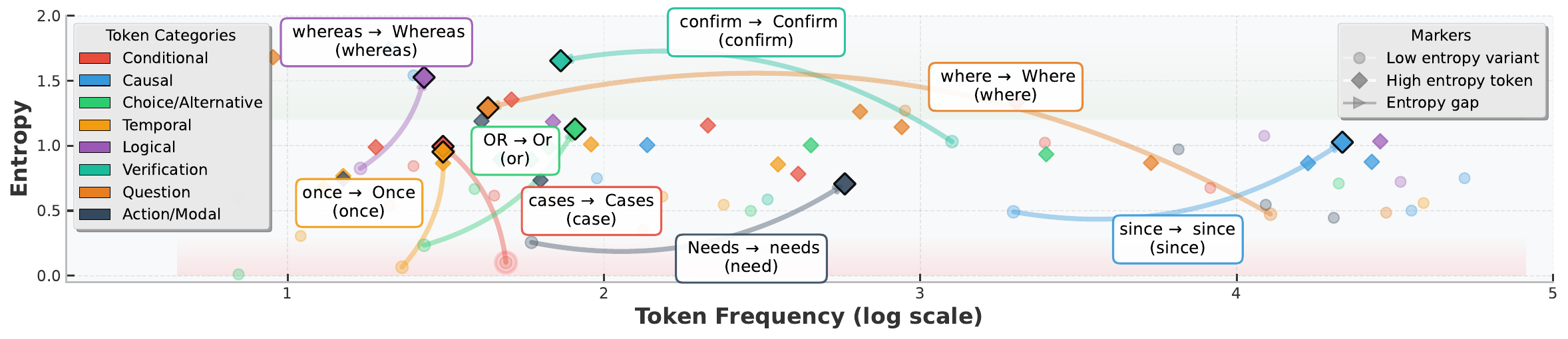}
    \caption{Dual-Entropy Token Frequency-Entropy Landscape}
    \label{fig:Dual_Entropy_Phenomenon2}
\end{figure*}
\begin{figure*}[!t]
    \centering
    \begin{subfigure}{0.26\textwidth}
        \centering
        \includegraphics[width=3.5cm,height=3.4cm]{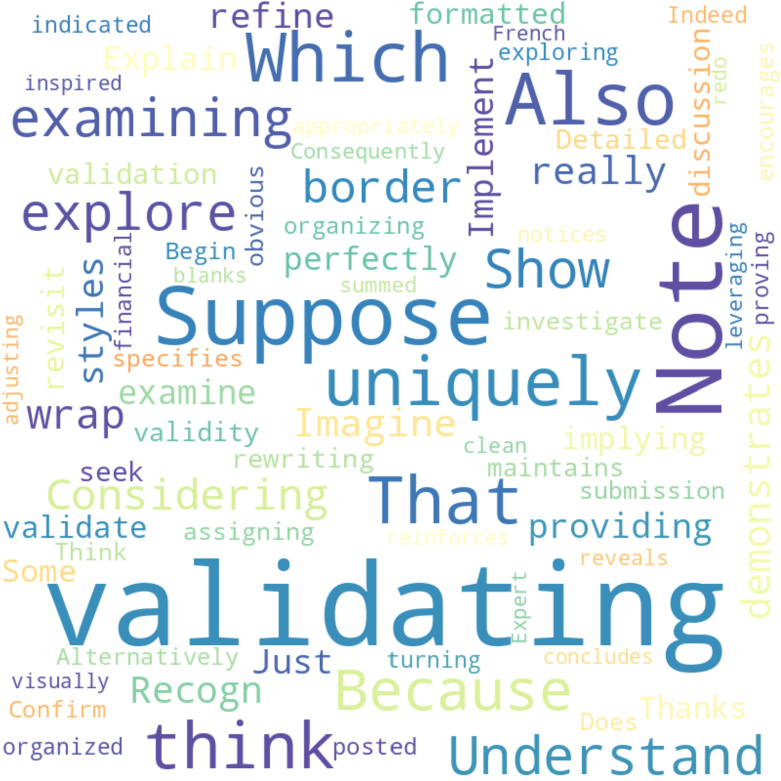}
        \caption{Top100 High Entropy Tokens}
        \label{fig:token_distributions2}
    \end{subfigure}
    \begin{subfigure}{0.21\textwidth}
        \centering
        \includegraphics[width=3.6cm,height=3.4cm]{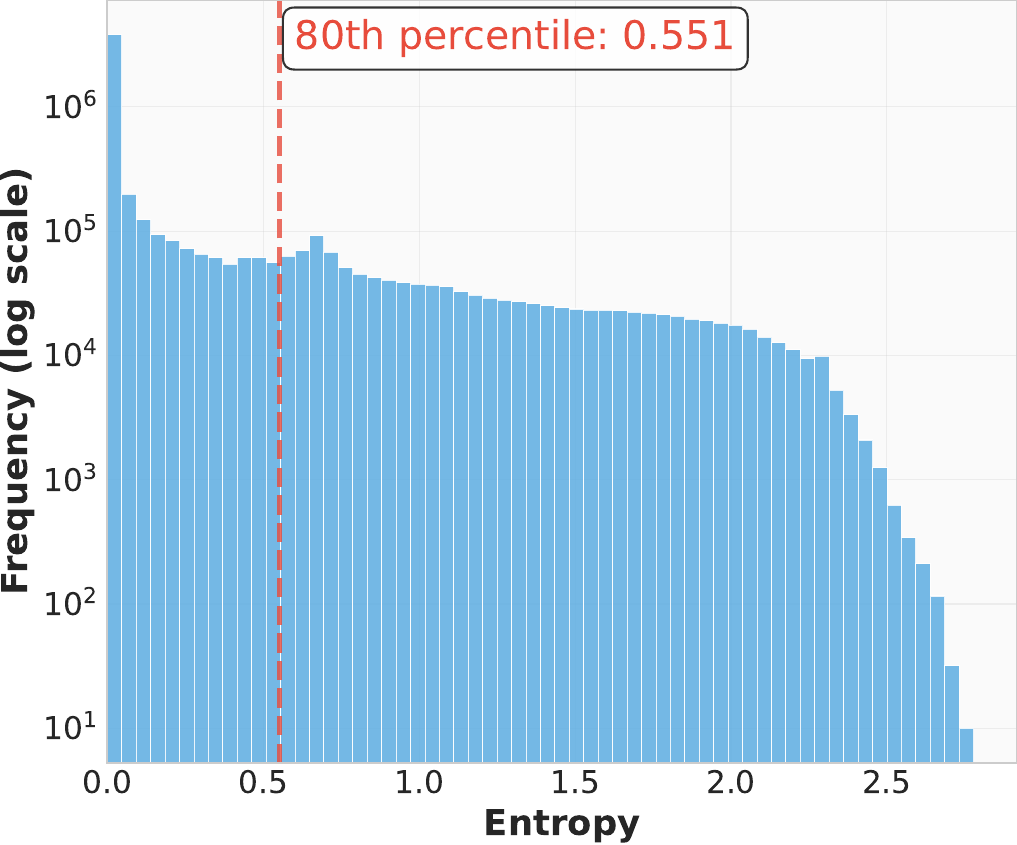}
        \caption{Entropy Distribution}
        \label{fig:token_distributions1}
    \end{subfigure}
    \hfill
    \hfill
    \begin{subfigure}{0.22\textwidth}
        \centering
        \includegraphics[width=3.6cm,height=3.4cm]{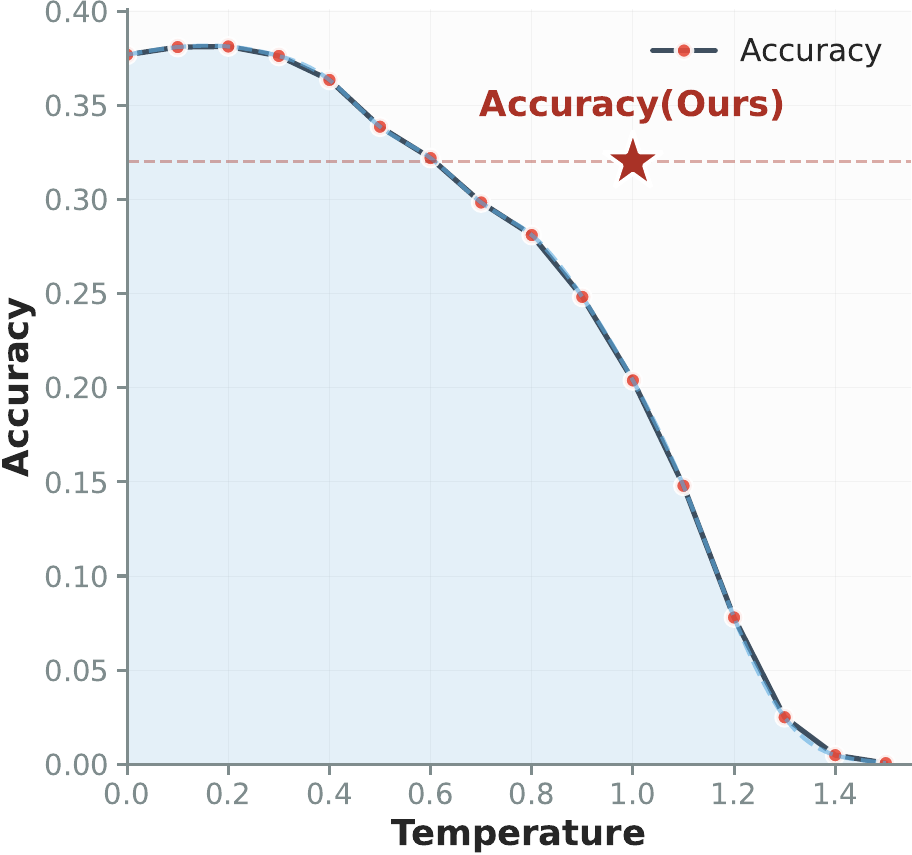}
        \caption{Accuracy}
        \label{fig:token_distributions3}
    \end{subfigure}
    \hfill
    \begin{subfigure}{0.26\textwidth}
        \centering
        \includegraphics[width=4.0cm,height=3.4cm]{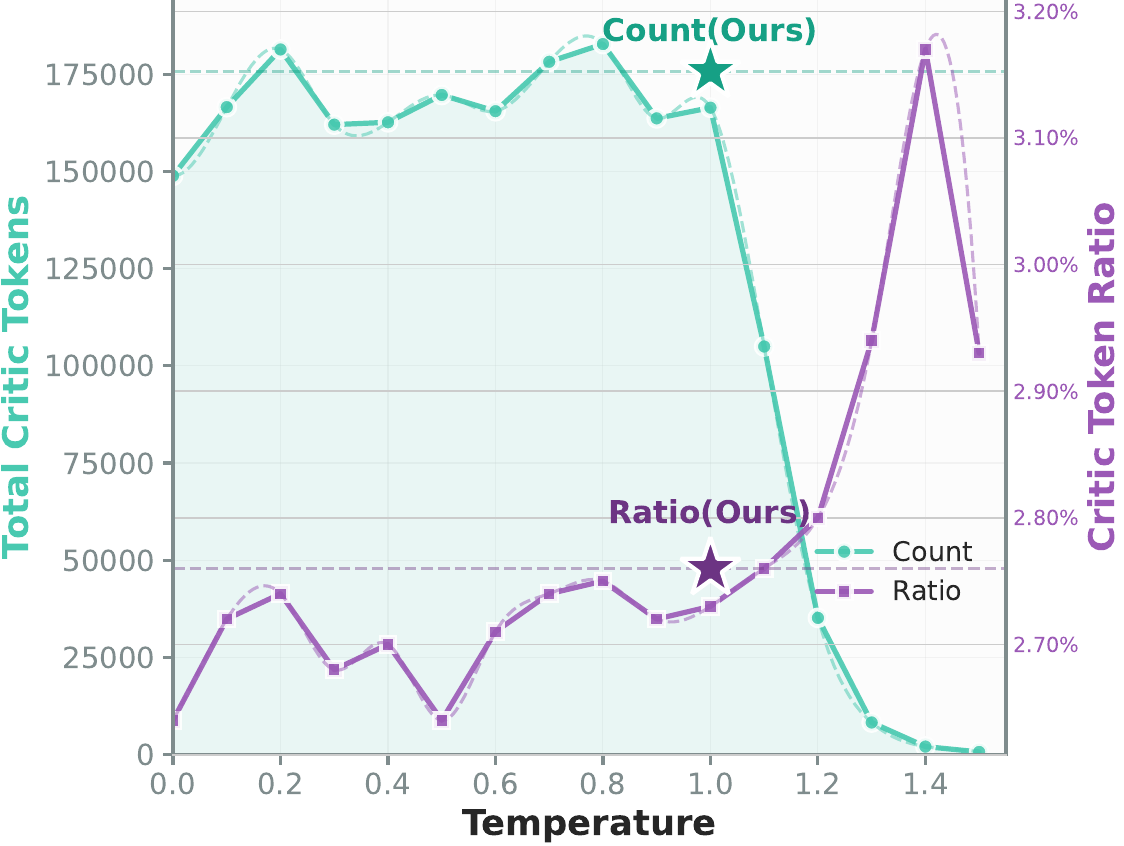}
        \caption{Critic Token Counts}
    \label{fig:token_distributions4}
    \end{subfigure}
    \caption{Token entropy characteristics and temperature effects on critical tokens.}
    \label{fig:token_distributions}
\end{figure*}

\begin{itemize}[itemsep=7pt, topsep=3pt, parsep=0pt]
\item \textbf{Adaptive Temperature Sampling}:  We dynamically adjust sampling temperature based on token entropy—increasing it for high-entropy tokens to promote exploration while reducing it for low-entropy tokens to maintain coherence. This resolves the accuracy-exploration trade-off in training data.
\item \textbf{Token-Level Group Average Advantage Estimation}: We estimate advantages at the token level across the group, respecting intra-sequence heterogeneity. This also ensures balanced optimization between positive and negative samples while preserving DAPO's token mean loss benefits for long-sequence learning.
\item \textbf{Differential Advantage Redistribution}: We leverage entropy and importance ratios to modulate advantages—amplifying advantages for high-entropy tokens with extreme ratios while suppressing low-entropy ones near unity, enabling fine-grained attribution aligned with each token's optimization needs.
\item \textbf{Asymmetric Adaptive Clipping}: We implement reversed asymmetric clipping boundaries. Low-entropy tokens receive expanded left boundaries to enable aggressive noise suppression, while high-entropy tokens receive expanded right boundaries to facilitate exploration at critical decision points.
\end{itemize}

HAPO realizes fine-grained token-level heterogeneous optimization based on each token's continuous entropy signal. Our experiments demonstrate that this systematic, continuous treatment yields substantial improvements on mathematical reasoning, code, and logic tasks across multiple models.

\section{Background}
\label{Background}
Given a language model $\pi_\theta$. States consist of the prompt $q$ and previously generated tokens $o_{<t} = (o_1, o_2, ..., o_{t-1})$. Actions are tokens $o_t$ from vocabulary $\mathcal{V}$. Rewards $R(o)$ are typically provided at the sequence level for the complete response $o$.

\textbf{Group Relative Policy Optimization.} GRPO~\citep{shao2024deepseekmathpushinglimitsmathematical} removes value function by employing group normalization. For each prompt $q$, GRPO samples $G$ responses $\{o_i\}_{i=1}^G$:
{
\begin{multline}
\mathcal{L}^{\text{GRPO}} = \mathbb{E}_{(q,o)} \Bigg[\frac{1}{G} \sum_{i=1}^{G} \frac{1}{|o_i|} \sum_{t=1}^{|o_i|} \min \Big( r_{i,t} {A}_{i,t},\\
\text{clip}(r_{i,t}, 1-\epsilon, 1+\epsilon) {A}_{i,t} \Big) - \beta D_{\text{KL}}\left[\pi_\theta \| \pi_{\text{ref}}\right] \Bigg]
\label{equ:grpo}
\end{multline}
}
The advantage is estimated at sequence-level within the group as ${A}_{i,t} = \frac{R_i - \text{mean}(\{R_j\}_{j=1}^G)}{\text{std}(\{R_j\}_{j=1}^G)}$.

\textbf{GRPO with Entropy Advantage}. Entropy Adv~\citep{cheng2025reasoningexplorationentropyperspective} augments the advantage function with an entropy-based term, encouraging longer and deeper reasoning chains:
{
\begin{multline}
\mathcal{L}^{\text{Ent}} = \mathbb{E}_{(q,o)} \Bigg[\frac{1}{G} \sum_{i=1}^{G} \frac{1}{|o_i|} \sum_{t=1}^{|o_i|} \min \Big( r_{i,t} A^{\text{s}}_{i,t},\\
\text{clip}(r_{i,t}, 1-\epsilon, 1+\epsilon) A^{\text{s}}_{i,t} \Big) - \beta D_{\text{KL}}\left[\pi_\theta \| \pi_{\text{ref}}\right] \Bigg]
\label{equ:entropy_grpo}
\end{multline}
}
where $A^{\text{s}}_{i,t} = A_{i,t} + \psi(\mathcal{H}_{t})$, combining the advantage $A_{i,t}$ with an entropy-based bonus $\psi(\mathcal{H}_{t})$.

\textbf{Decoupled
Clip and Dynamic Ampling Policy Optimization.} DAPO ~\citep{yu2025dapoopensourcellmreinforcement} modifies GRPO with token-mean loss—averaging over tokens rather than sequences—to preserve gradient contributions from longer sequences. It also introduces clip-higher with asymmetric bounds:
{
\begin{multline}
\mathcal{L}^{\text{DAPO}} = \mathbb{E}_{(q,o)} \left[ \frac{1}{\sum_{i=1}^G |o_i|} \sum_{i=1}^{G} \sum_{t=1}^{|o_i|} \min \left( r_{i,t} {A}_{i,t}, \right. \right. \\
\left. \left. \text{clip}(r_{i,t}, 1-\epsilon_{\text{low}}, 1+\epsilon_{\text{high}}) \hat{A}_{i,t} \right) \right]
\label{equ:dapo}
\end{multline}
}
The asymmetric bounds $\epsilon_{\text{high}} > \epsilon_{\text{low}}$ (typically $\epsilon_{\text{high}} = 0.28$, $\epsilon_{\text{low}} = 0.2$) allow larger probability increases for high-entropy token exploration.

\textbf{DAPO with Forking Tokens.} ~\citep{wang20258020rulehighentropyminority} demonstrates that a minority of high entropy tokens function as critical decision points. This motivates selective optimization focusing on high-entropy tokens:
{
\begin{multline}
\mathcal{L}^{\text{Ent}} = \mathbb{E}_{(q,o)} \Biggl[ \frac{1}{\sum_{i=1}^G |o_i|} \sum_{i=1}^{G} \sum_{t=1}^{|o_i|} \mathbb{I}[H_t \geq \tau_{\rho}] \cdot \\
\min\left( r_{i,t} A_{i,t}, \operatorname{clip}(r_{i,t}, 1-\epsilon_{\text{low}}, 1+\epsilon_{\text{high}}) A_{i,t} \right) \Biggr]
\end{multline}
where $H_t$ represents the entropy at position $t$, $\tau_{\rho}$ denotes the $\rho$-th percentile entropy threshold.
}
\begin{figure*}[!t]
    \centering
    \begin{subfigure}{0.24\textwidth}
        \centering
        \includegraphics[width=3.8cm,height=3.4cm]{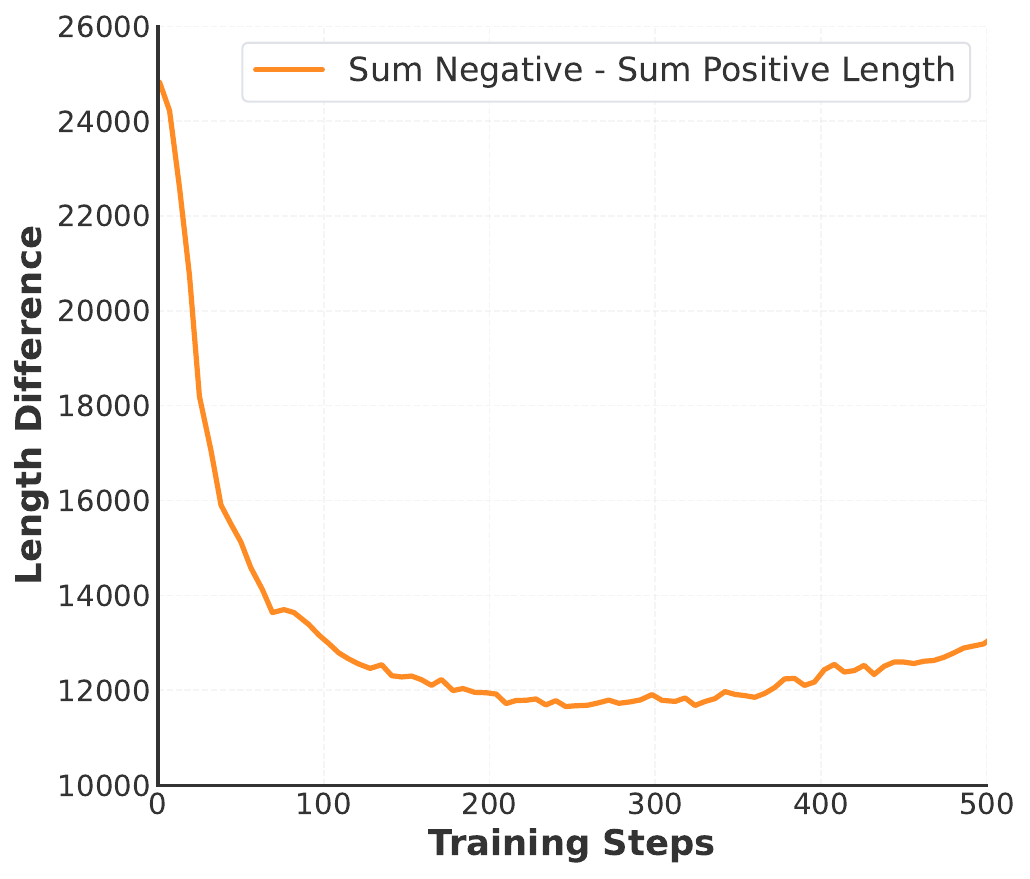}
        \caption{Length Differences}
        \label{fig:adv_analysis_part1}
    \end{subfigure}
    \hfill
    \begin{subfigure}{0.24\textwidth}
        \centering
        \includegraphics[width=3.8cm,height=3.4cm]
        {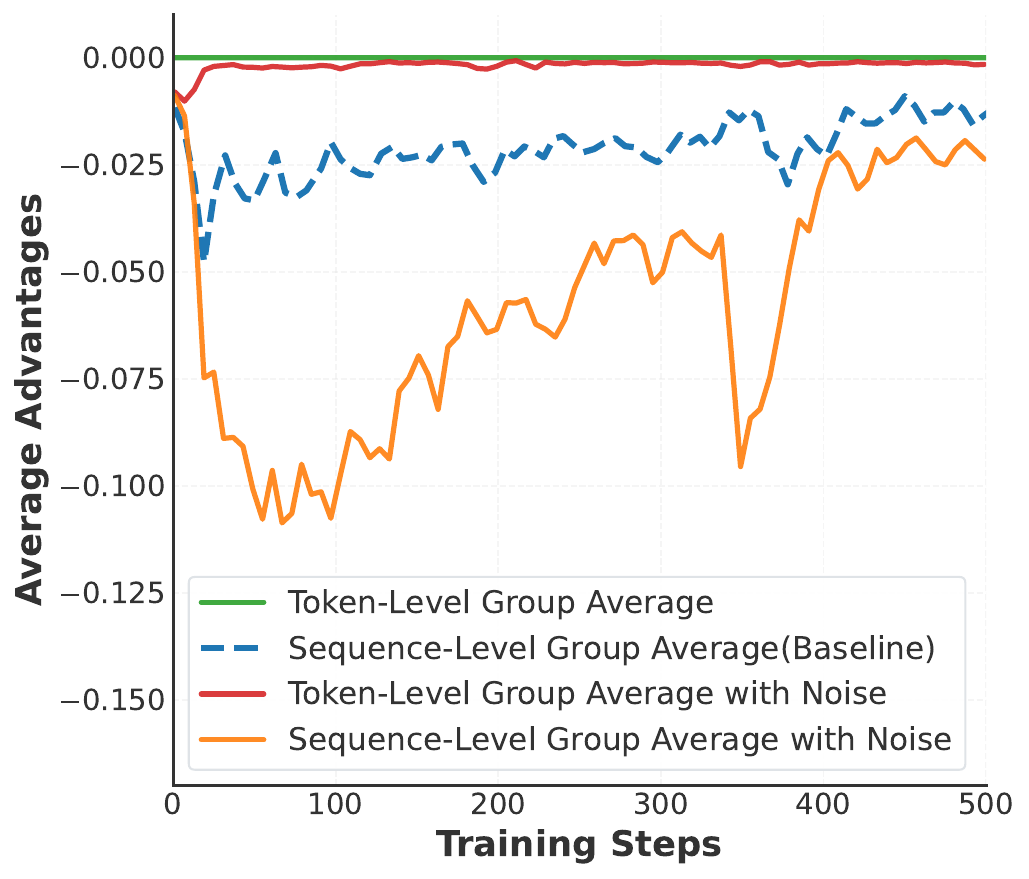}
        \caption{Average Advantage}
        \label{fig:adv_analysis_part2}
    \end{subfigure}
    \hfill
    \begin{subfigure}{0.24\textwidth}
        \centering
        \includegraphics[width=3.8cm,height=3.4cm]{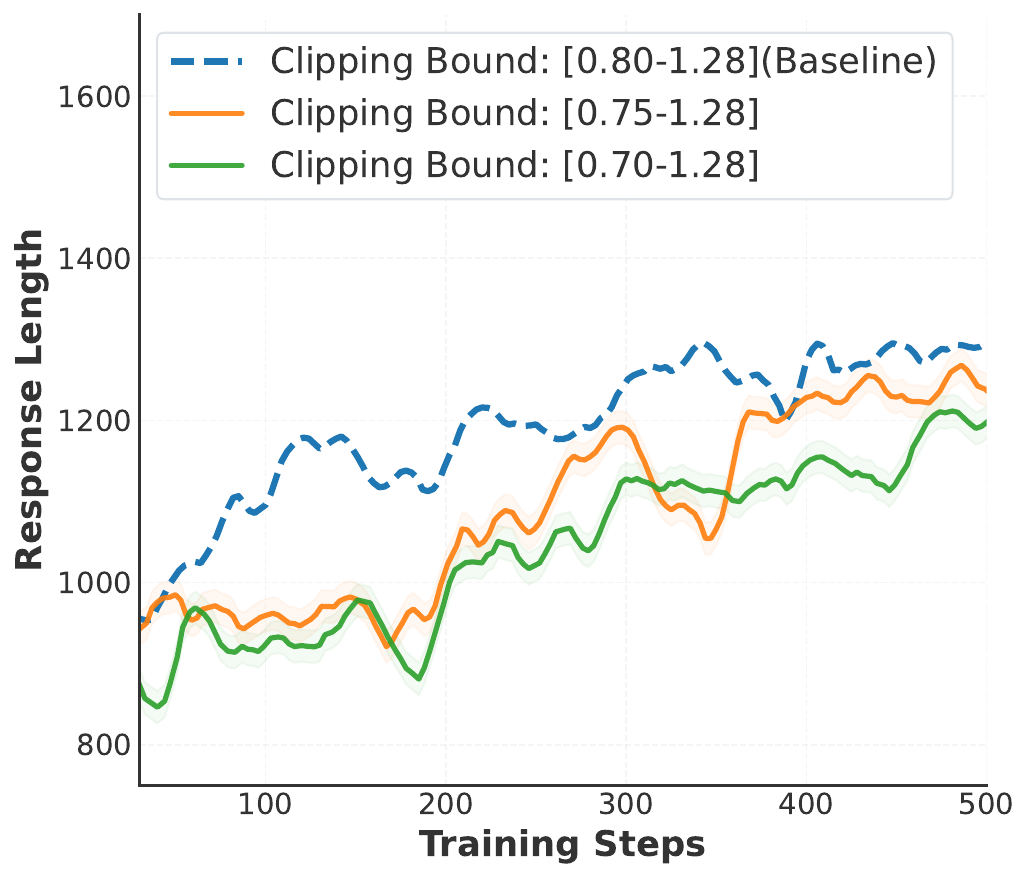}
        \caption{Response Length}
        \label{fig:left_analysis_part1}
    \end{subfigure}
    \hfill
    \begin{subfigure}{0.26\textwidth}
        \centering
        \includegraphics[width=3.8cm,height=3.4cm]{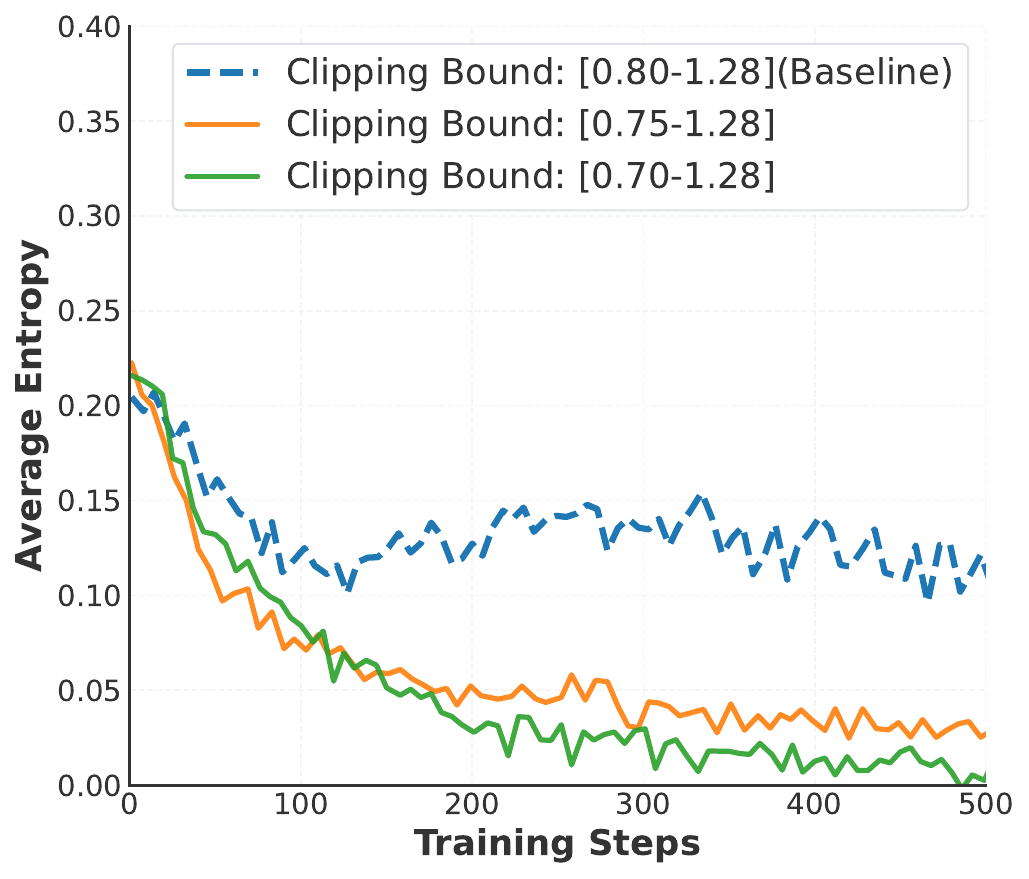}
        \caption{Average Entropy}
        \label{fig:left_analysis_part2}
    \end{subfigure}
    \caption{Bias of token-mean loss toward negative samples affects advantage and clipping.}
    \label{fig:adv_dynamic}
\end{figure*}

\begin{figure*}[!t]
    \centering
    \begin{subfigure}{0.67\textwidth}
        \centering
        \includegraphics[width=10.2cm,height=3.3cm]{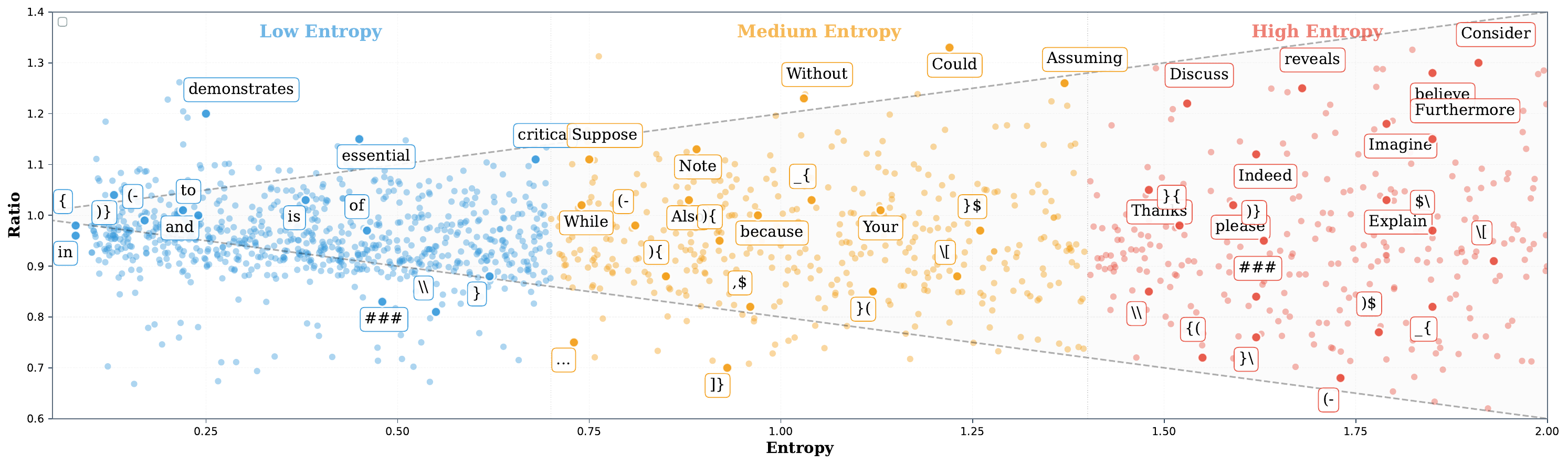}
        \caption{Token Distribution: Entropy-Dependent Ratio Spread}
        \label{fig:adv_ratio1}
    \end{subfigure}
    \hfill
    \begin{subfigure}{0.30\textwidth}
        \centering
        \includegraphics[width=3.9cm,height=3.3cm]{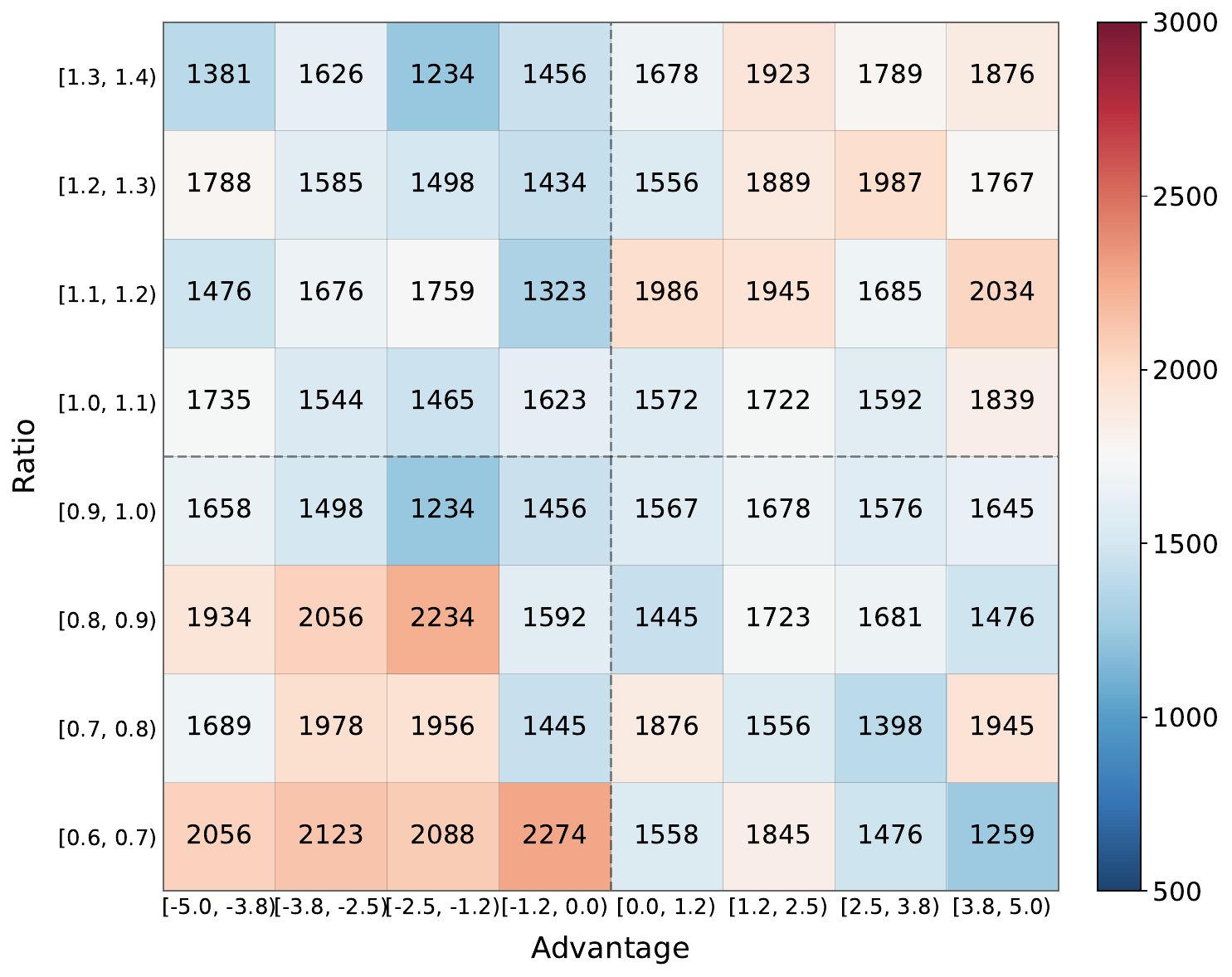}
        \caption{Ratio-Advantage Heatmap}
        \label{fig:adv_ratio2}
    \end{subfigure}
    \caption{Token update patterns analysis via importance ratios}
    \label{fig:Ratio_Update_Analysis}
\end{figure*}
\section{Token Heterogeneity in RLHF}
\label{sec:Token Heterogeneity}
\subsection{Why Heterogeneity Is Necessary}
\textbf{The Dual-Entropy Phenomenon.}
We first analyzed entropy patterns during training. As shown in Figures~ \ref{fig:Dual_Entropy_Phenomenon2}, we discovered "Dual-Entropy Phenomenon": high-entropy tokens frequently have "twin siblings" in low-entropy region—tokens with same word stems but drastically different entropy. We show more visualizations in Appendix ~\ref{app:dual_entropy}. 

This pattern spans multiple semantic categories and frequency ranges, explaining why low-entropy tokens are crucial: they contain important reasoning tokens that differ from high-entropy ones only in syntactic context. This reveals limitations of entropy-based binary discretization and uniform treatment, necessitating a nuanced approach that preserves low-entropy tokens' stabilizing influence without letting them dominate the learning signal.

\begin{figure*}[!t]
    \centering
    \begin{subfigure}{0.24\textwidth}
        \centering
        \includegraphics[width=3.8cm,height=3.4cm]{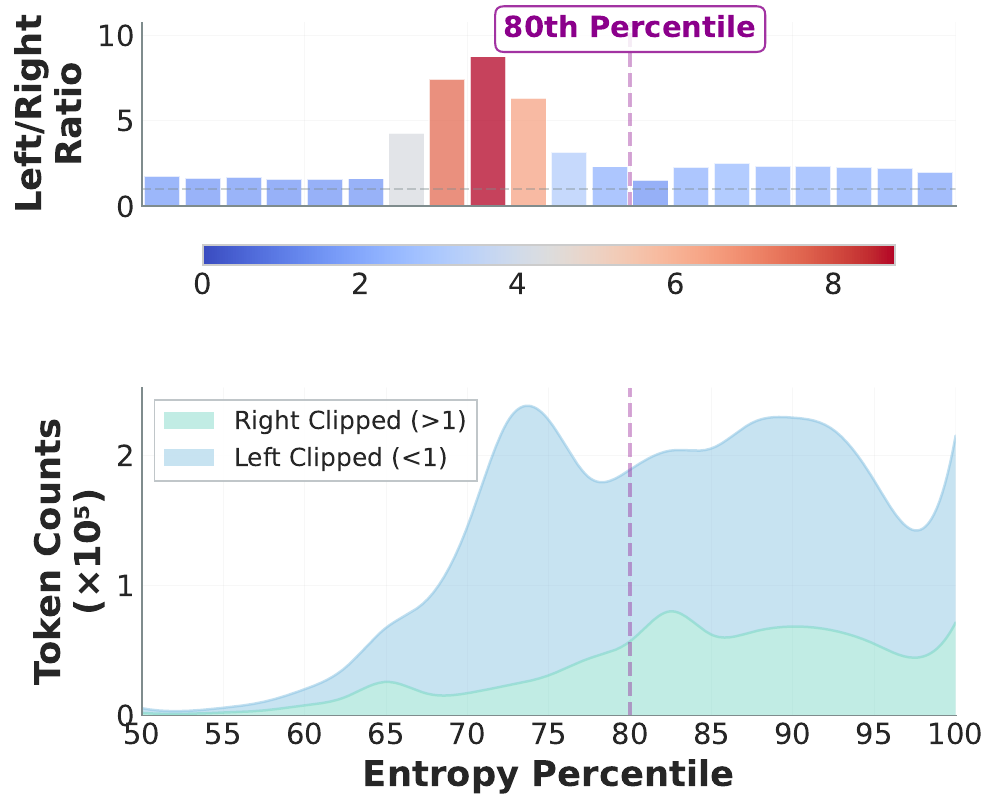}
        \caption{Entropy-Clip Patterns}
        \label{fig:clip_analysis_part1}
    \end{subfigure}
    \hfill
    \begin{subfigure}{0.24\textwidth}
        \centering
        \includegraphics[width=3.6cm,height=3.4cm]{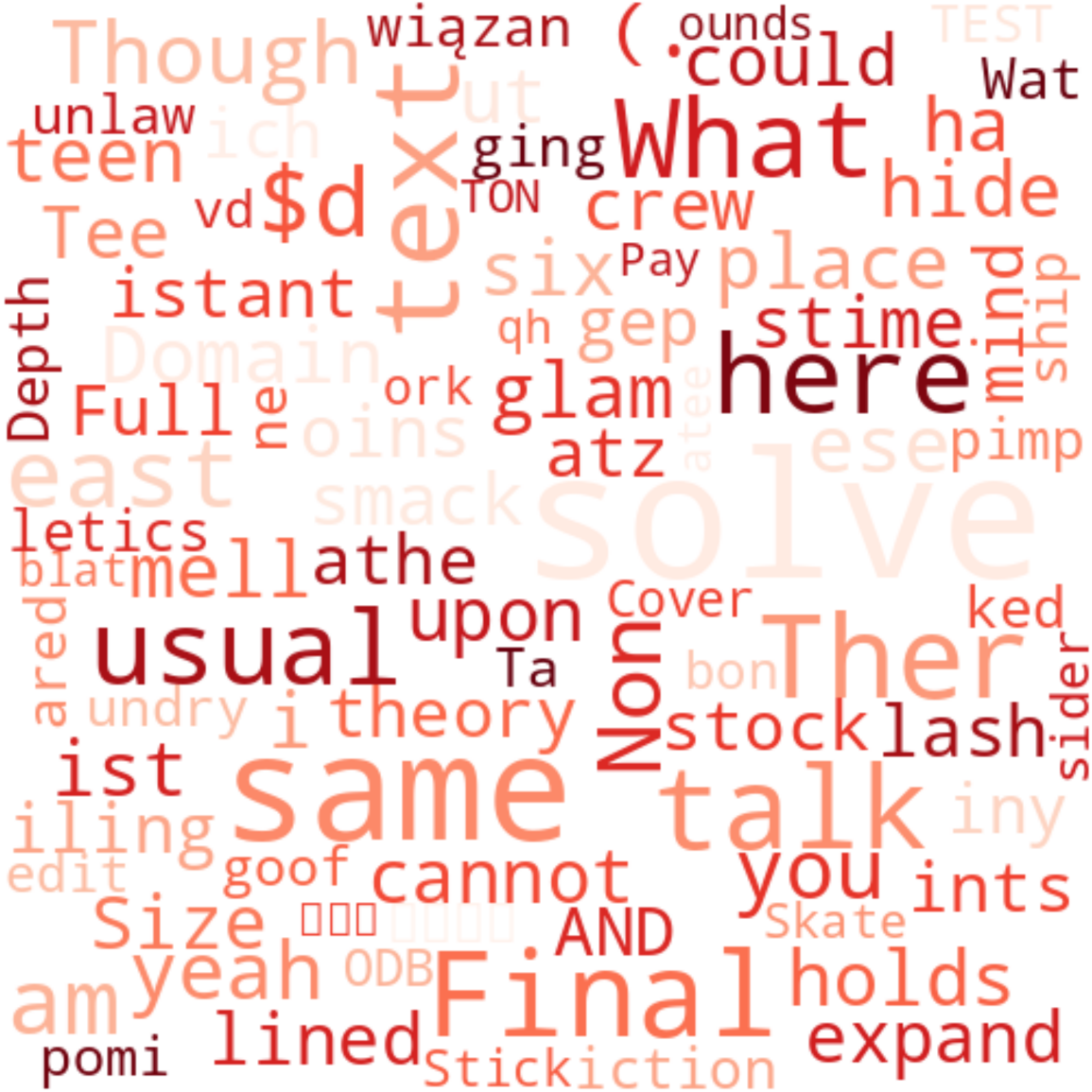}
        \caption{Right-Clipped}
        \label{fig:clip_analysis_part3}
    \end{subfigure}
    \hfill
    \begin{subfigure}{0.26\textwidth}
        \centering
        \includegraphics[width=3.6cm,height=3.4cm]{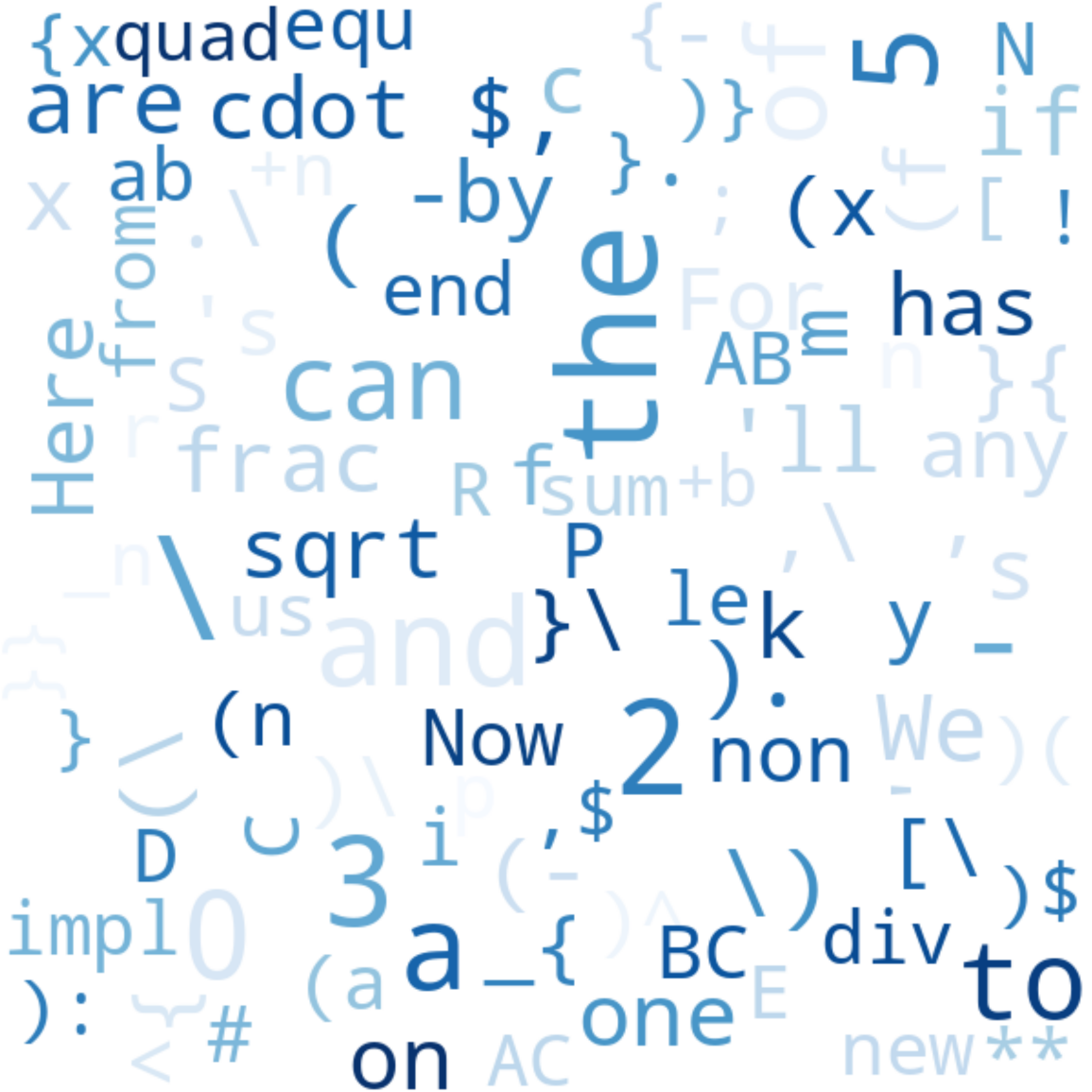}
        \caption{Left-Clipped}
        \label{fig:clip_analysis_part4}
    \end{subfigure}
    \hfill
    \begin{subfigure}{0.24\textwidth}
        \centering
        \includegraphics[width=3.9cm,height=3.4cm]{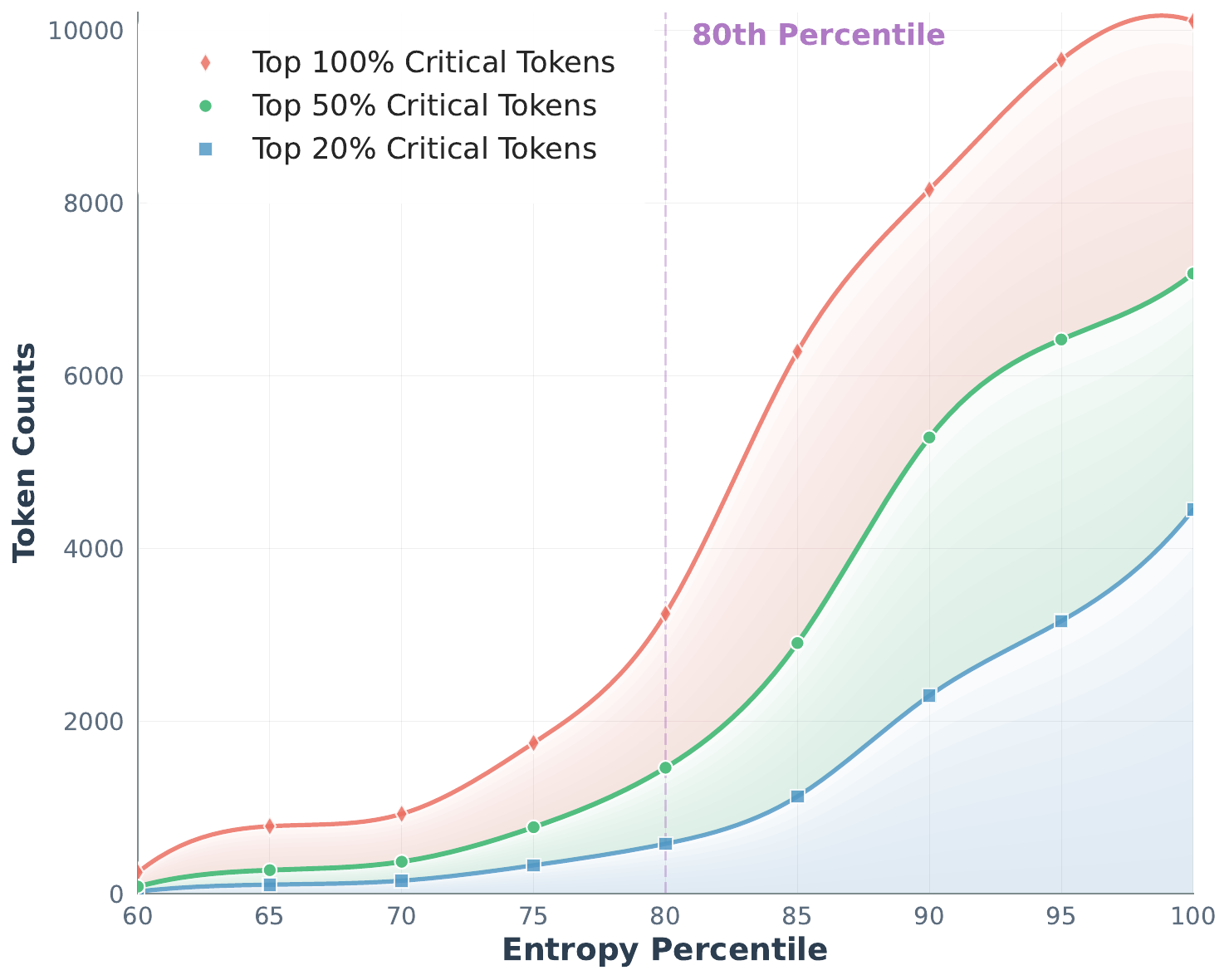}
        \caption{Critical Token Clipped}
        \label{fig:clip_analysis_part2}
    \end{subfigure}
    \caption{Token-Level clipping patterns and word cloud visualizations}
    \label{fig:Clipped_Token_Properties}
\end{figure*}

\subsection{Rollout Generation: Exploration-Exploitation Imbalance}
\label{Rollout_Generation}
\textbf{Characterizing Tokens in Rollout Generation.}
We then analyze token distributions and occurrence frequencies in Figure~\ref{fig:token_distributions2},~\ref{fig:token_distributions1}. We find that high-entropy tokens are rare. These tokens have low sampling probabilities but predominantly represent critical decision points. This creates a paradox: the tokens most valuable for learning are systematically under-sampled during standard rollout.

\textbf{Temperature Effects on Critic Tokens.}
To address this, we investigate temperature's role in generating critical tokens. We evaluate rollout across various temperatures. Figure~\ref{fig:token_distributions3},~\ref{fig:token_distributions4} reveal that while accuracy decreases with temperature, critical token occurrence increases before dropping at extremes. This demonstrates a fundamental trade-off: low temperatures preserve accuracy but suppress critical tokens, while high temperatures generate more meaningful tokens but reduce accuracy.
\subsection{Advantage Calculation: Reward Attribution Granularity}
\label{sec:Advantage Calculation}
\textbf{Limitations of Sequence-Level Advantage Estimation.}
Current methods adopt GRPO's sequence-level normalization to estimate advantages—using a single reward to represent an entire sequence for estimation, disregarding length differences across sequences. Figure~\ref{fig:adv_analysis_part1} reveals that negative samples are significantly longer than positive ones. GRPO divides loss by sequence length (Equation \ref{equ:grpo}) to balance contributions but sacrifices individual token gradients. DAPO improves upon this by dividing by total tokens (Equation \ref{equ:dapo}) to preserve token-level gradients—a natural approach for handling token heterogeneity. However, this creates a conflict: sequence-level advantages estimation with token-mean loss causes longer negative samples to dominate gradient, as shown in Figure~\ref{fig:adv_analysis_part2}.

While DAPO's token-mean loss achieves better heterogeneity, sequence-level advantage estimation introduce negative length bias that impacts token-level optimization. First, when minor perturbations are applied to advantages, the ignorance of length in sequence-level advantage estimation amplifies these fluctuations into substantial deviations, causing training instability as shown in Figure~\ref{fig:adv_analysis_part2}. Second, this bias conflicts with strategies like Archer, which relaxes left clipping bounds for high-entropy tokens. As shown in Figure~\ref{fig:left_analysis_part1} and Figure~\ref{fig:left_analysis_part2}, negative advantage dominance combined with relaxed left clipping bounds causes rapid probability degradation for high-entropy critical tokens, leading to entropy collapse and reduced response lengths.

\textbf{Analyzing Token Update Patterns.}
Beyond sequence-level advantage ignoring token heterogeneity, another limitation is that all tokens within a sequence receive same advantages. We conduct a fine-grained analysis of advantage dynamics during training. Specifically, we introduce importance ratios $r_{i,t} = \pi_\theta(a_{i,t}|s_{i,t}) / \pi_{\text{ref}}(a_{i,t}|s_{i,t})$, which reflect model's existing optimization state and actual update needs for each token.

Figure~\ref{fig:adv_ratio1} shows high-entropy tokens have broader ratio distributions, with most tokens deviating from 1.0 as the model preferentially updates them. By analyzing the distribution of top 20\% high-entropy tokens across different advantage and ratio values in Figure~\ref{fig:adv_ratio2}, we observe a strong correlation between advantages and ratios. Meanwhile, we find that low-entropy tokens' ratios cluster around 1.0, suggesting no clear update direction. The token examples in Figure~\ref{fig:adv_ratio1} demonstrate that tokens with large ratio deviations are those requiring the most optimization—either critical tokens or redundant tokens.

These findings demonstrate that tokens with similar entropy often have vastly different optimization needs. This insight exposes critical flaws in prior works that leverage entropy to reshape advantages. Methods like Entropy Adv incorporate entropy as an intrinsic reward added to the advantage (Eq.~\ref{equ:entropy_grpo}), which boosts all high-entropy tokens. For high-entropy tokens with negative advantages, this entropy reward offsets the negative advantage of erroneous tokens, dampening penalties for errors and thereby weakening the training signal.

\subsection{Clipping Loss Computation: Clipping Constraint} 
\textbf{Analyzing Clipped Token Properties.}
Figures~\ref{fig:Clipped_Token_Properties} reveal distinct clipping patterns: low-entropy tokens, mostly format and useless code tokens, are frequently left-clipped, while high-entropy tokens including critical reasoning tokens are frequently right-clipped. Figure~\ref{fig:clip_analysis_part2} shows that even with DAPO's clip higher, high-entropy tokens remain frequently clipped, with frequency increasing with entropy. Existing clipping strategies prematurely exclude these tokens from gradient updates. Uniform clipping constrains both useless low-entropy tokens that need probability reduction and critical high-entropy tokens that require exploration. DAPO's static high clipping bound is inadequate for adapting to token-specific entropy and update requirements. Furthermore, its restrictive left bound hinders the model from aggressively suppressing useless low-entropy tokens.
\section{Heterogeneous Adaptive Policy Optimization}
\label{method}
Based on the challenges in Section \ref{sec:Token Heterogeneity}, we propose HAPO that optimizes the RL pipeline according to token heterogeneity. We introduce our core principles and formalize how entropy serves as a continuous regulator, enabling smooth transitions between different tokens while maintaining simplicity.
\subsection{Entropy-Based Continuous Regulation}
As we use entropy as a continuous indicator, we first prepare the entropy to obtain a unified regulation signal. Following ~\citep{wang20258020rulehighentropyminority}, we normalize entropy within each batch and apply asymmetric scaling to map values to $[-1, 1]$:
\begin{equation}
h_{i,t} = \frac{\log(H_{i,t}) - Q_{\rho}(\log(H))}{\sigma(\log(H))}
\end{equation}
\begin{equation}
\tilde{h}_{i,t} = \begin{cases}
h_{i,t} / h_{\max} & \text{if } h_{i,t} > 0 \\
-h_{i,t} / |h_{\min}| & \text{if } h_{i,t} \leq 0
\end{cases}
\end{equation}
where $H_{i,t}$ is the entropy at ${i,t}$, $Q_{\rho}$ denotes the $\rho$-th quantile serving as the boundary between high and low entropy tokens, and $\sigma$ is the standard deviation. 

We apply logarithmic smoothing to handle the significant variance in entropy distribution. The asymmetric scaling ensures balanced regulation across both positive and negative ranges.

With this normalized entropy $\tilde{h}_{i,t}$, we implement adaptive modulation across multiple components:
\begin{equation}
\text{Parameter}_{i,t} = \text{Parameter}_{\text{base}} \cdot f(\tilde{h}_{i,t})
\label{equ:format_equation}
\end{equation}
where $f(\cdot)$ is designed for each specific parameters. 

Equation~\ref{equ:format_equation} serves as a meta-equation for our adaptive strategy. We apply this logic by substituting the parameters with sampling temperature, advantage estimates, and clipping bounds.
\subsection{Adaptive Temperature Sampling}
Now that we understand temperature's impact on token generation and accuracy, a balanced strategy is to introduce dynamic temperature. We apply high temperature to high-entropy tokens to encourage exploration while using low temperature for low-entropy ones to maintain coherence. Since rollout proceeds token-by-token without sequence-level context, we maintain the formulation of Equation~\ref{equ:format_equation} but dynamically adjust temperature based on current state, computing normalization on-the-fly:
\begin{equation}
T_{i,t} = T_{\text{base}} \cdot \left(1 + \frac{\log(H_{i,t}) - {\hat\rho_{\log(H)}}}{\hat\sigma_{\log(H)}} \cdot \tau \right)
\end{equation}
$\hat\rho_{\log(H)}$ is the $\rho$-th quantile of entropy and serves as the direction of temperature adjustments. $\hat\sigma_{\log(H)}$ is the variance, and $\tau$ determines the maximum bounds of temperature adjustment. We compute $\hat\rho_{\log(H)}$ and $\hat\sigma_{\log(H)}$ by leveraging token statistics from previous step to capture more comprehensive information, rather than relying on sequence-level normalization.

\begin{table*}[!t]
    \small
    \setlength{\extrarowheight}{3pt}
    \setlength{\tabcolsep}{6pt}
    \renewcommand{\arraystretch}{0.85}
    \centering
        \caption{Comparison between \emph{HAPO} and existing methods on Math benchmarks. The models below are from 1000 steps of training. The best results are indicated by \textbf{boldface}.}
        \label{tab: qwen_math_results}
    \resizebox{1.0\textwidth}{!}{%
    \begin{tabular}{@{}ccccccccc@{}}
    \toprule
    \textbf{Model} & \textbf{Method} & \textbf{Avg (1590)} & AIME24(30) & AIME25(30) & AMC(83) & Math(500) & OlympiadBench(675) & Minerva (272) \\
    \cmidrule(r){1-1} \cmidrule(lr){2-2} \cmidrule(l){3-9}
    
     \multirow{5}{*}{\begin{tabular}[c]{@{}l@{}}Qwen2.5-Math-1.5B\end{tabular}}

     & Vanilla GRPO & 36.83 & 20.12 & 15.67 & 71.23 & 65.91 & 29.84 & 18.23 \\
    
     & Vanilla DAPO & 38.34 & 21.73 & 17.11 & 72.85 & \textbf{67.33} & 31.47 & 19.52 \\
     
     & DAPO w/ Forking Tokens  & 38.83 & 22.51 & 18.56 & 74.80 & 65.54 & 32.81 & 18.74 \\
     
     & Archer & 37.01 & 19.62 & 17.24 & 72.17 & 64.21 & 29.98 & 19.14 \\

     & Entropy Env & 37.45 & 19.45 & 18.02 & 72.88 & 65.13 & 29.56 & 19.67 \\
    
     & EDGE-GRPO & 38.79 & 22.88 & 17.86 & 74.48 & 66.94 & 30.67 & 19.93 \\
     
     & HAPO(ours) & \textbf{40.62} & \textbf{25.33} & \textbf{20.12} & \textbf{75.69} & 66.83 & \textbf{33.86} & \textbf{21.91}\\
    \midrule
    
    \multirow{5}{*}{\begin{tabular}[c]{@{}l@{}}Qwen2.5-Math-7B\end{tabular}}

    & Vanilla GRPO & 45.26 & 35.67 & 18.91 & 80.14 & 74.28 & 32.76 & 29.85 \\
    
     & Vanilla DAPO & 46.97 & 37.24 & 20.24 & 81.99 & 76.72 & 34.35 & 31.28 \\
     
     & DAPO w/ Forking Tokens  & 47.43 & 38.45 & 21.90 & 80.55 & 75.38 & 35.52 & 32.79 \\
     
     & Archer & 45.63 & 35.25 & 19.19 & 82.53 & 72.16 & 33.04 & 31.61 \\

     & Entropy Env & 46.15 & 36.82 & 18.95 & 82.94 & 73.05 & 33.68 & 31.43 \\
    
     & EDGE-GRPO & 45.89 & 38.85 & 21.83 & 83.23 & 69.78 & 32.43 & 29.24 \\
     
     & HAPO(ours) & \textbf{50.04} & \textbf{41.31} & \textbf{24.34} & \textbf{85.47} & \textbf{78.45} & \textbf{36.94} & \textbf{33.73} \\

    \midrule
    
    \multirow{5}{*}{\begin{tabular}[c]{@{}l@{}}Qwen3-8B\end{tabular}}

    & Vanilla GRPO & 48.15 & 33.91 & 23.56 & 76.84 & 81.17 & 42.93 & 30.47 \\
    
     & Vanilla DAPO & 50.00 & 35.84 & 25.24 & 78.37 & 83.54 & 44.87 & 32.13 \\
     
     & DAPO w/ Forking Tokens  & 50.21 & 36.22 & 25.85 & 80.26 & 82.75 & \textbf{46.27} & 29.91 \\
     
     & Archer & 49.17 & 34.76 & 23.33 & 78.47 & 82.57 & 42.69 & 33.22 \\

    & Entropy Env & 49.60 & 34.45 & 24.12 & 80.05 & 83.12 & 42.88 & 32.95 \\
    
     & EDGE-GRPO & 50.06 & 35.15 & 24.12 & 80.47 & 83.33 & 43.35 & 33.91 \\
     
     & HAPO(ours) & \textbf{51.97} & \textbf{39.01} & \textbf{26.83} & \textbf{81.77} & \textbf{84.23} & 45.26 & \textbf{34.74} \\

    \midrule
    
    \multirow{5}{*}{\begin{tabular}[c]{@{}l@{}}Qwen3-14B\end{tabular}}

     & Vanilla GRPO & 55.89 & 42.36 & 35.78 & 83.91 & 85.63 & 50.14 & 37.52 \\
    
     & Vanilla DAPO & 58.06 & 44.82 & 37.53 & 85.52 & 88.25 & 52.28 & 39.97\\
     
     & DAPO w/ Forking Tokens & 60.00 & 46.71 & 39.12 & 87.31 & 88.90 & \textbf{56.85} & 40.84 \\

     & Archer & 58.88 & 46.28 & 38.68 & 86.73 & 87.76 & 53.19 & 40.62 \\

     & Entropy Env & 59.27 & 46.35 & 39.22 & 87.04 & 88.35 & 54.12 & 40.56 \\
    
     & EDGE-GRPO & 59.16 & 46.94 & 39.01 & 87.15 & 89.12 & 52.45 & 40.31 \\
     
     & HAPO(ours) & \textbf{62.09} & \textbf{49.74} & \textbf{41.71} & \textbf{89.66} & \textbf{92.47} & 56.43 & \textbf{42.53}\\
     
    \bottomrule
    \end{tabular}
    }
\end{table*}
\begin{table*}[!t]
    \small
    \setlength{\extrarowheight}{3pt}
    \setlength{\tabcolsep}{6pt}
    \renewcommand{\arraystretch}{0.8}
    \centering
        \caption{Comparison between \emph{HAPO} and existing methods on LLaMA series models.}
        \label{tab: llama_results}
    \resizebox{1.0\textwidth}{!}{%
    \begin{tabular}{@{}ccccccccc@{}}
    \toprule
    \textbf{Model} & \textbf{Method} & \textbf{Avg (1590)} & AIME24(30) & AIME25(30) & AMC(83) & Math(500) & OlympiadBench(675) & Minerva (272) \\
    \cmidrule(r){1-1} \cmidrule(lr){2-2} \cmidrule(l){3-9}

    \multirow{3}{*}{\begin{tabular}[c]{@{}l@{}}LLaMA3.2-3B Instruct\end{tabular}}
     & Vanilla GRPO & 20.97 & 11.67 & 0.00 & 49.28 & 38.62 & 14.35 & 11.94 \\
     
     & Vanilla DAPO & 22.99 & 13.93 & 0.00 & 52.11 & 41.95 & 16.63 & 13.32\\
     
     & DAPO w/ Forking Tokens  & 23.86 & 12.45 & 0.00 & 55.96 & 43.34 & 16.89 & 14.54 \\
     
     & HAPO(ours) & \textbf{27.42} & \textbf{17.50} & \textbf{8.00} & \textbf{58.74} & \textbf{44.86} & \textbf{19.37} & \textbf{16.05}\\

    \midrule
    
    \multirow{3}{*}{\begin{tabular}[c]{@{}l@{}}LLaMA3.1-8B-Instruct \end{tabular}}
     & Vanilla GRPO & 22.07 & 10.83 & 0.00 & 54.16 & 40.78 & 13.81 & 12.89 \\
     
     & Vanilla DAPO & 24.21 & 13.33 & 0.00 & 57.11 & 44.05 & 16.03 & 14.76 \\
     
     & DAPO w/ Forking Tokens  & 25.30 & 14.50 & 0.00 & 58.46 & 46.55 & 15.89 & 16.39 \\
     
     & HAPO(ours) & \textbf{27.06} & \textbf{19.17} & \textbf{1.25} & \textbf{59.32} & \textbf{47.83} & \textbf{17.66} & \textbf{17.11} \\
     
    \bottomrule
    \end{tabular}
    }
\end{table*}

\subsection{Token-Level Group Average Advantage Estimation}
As DAPO computes loss through token-mean, a more compatible approach for advantage estimation is token-level averaging. We estimate advantages across all tokens in the group, distributing rewards to individual tokens before normalization:
\begin{equation}
A_{i,t} = \frac{a_{i,t} - \mu_{\text{tok}}}{\sigma_{\text{tok}}}
\end{equation}
where $a_{i,t} = r_i \in \{0,1\}$ is the token-level reward inherited from sequence $i$, $\mu_{\text{tok}} = \frac{1}{|\mathcal{T}|} \sum_{(i,t) \in \mathcal{T}} a_{i,t}$ is the mean across all tokens, $\sigma_{\text{tok}} = \sqrt{\frac{1}{|\mathcal{T}|} \sum_{(i,t) \in \mathcal{T}} (a_{i,t} - \mu_{\text{tok}})^2}$ is the standard deviation across all tokens, $\mathcal{T} = \{(i,t)\}$ represents all tokens across sequences in the group.

The key distinction from sequence-level advantage estimation lies in granularity. Instead of using a single reward per sequence, we assign rewards to individual tokens then involve all tokens in advantage estimation. This ensures $\sum_{(i,t) \in \mathcal{T}} \hat{A}_{i,t} = 0$, resolving the gradient bias in token-mean loss. More importantly, our method preserves length-dependent gradient scaling within reward categories—longer sequences contribute larger gradients for complex reasoning. This simultaneously maintains GRPO's unbiased gradient contributions and DAPO's token-level granularity advantages.

\subsection{Differential Advantage Redistribution}

Our core idea remains amplifying advantages for high-entropy tokens while reducing them for low-entropy tokens. Different from existing methods, we additionally use importance ratios for regulation, ensuring updates based on each token's actual conditions to achieve token-level control. 

To implement this, we define a neutral zone $[\gamma_L, \gamma_U]$, typically centered around ratio = 1. High-entropy tokens receive amplification only when their importance ratio falls outside the neutral zone, reflecting clear update trends. We enhance them accordingly. For low-entropy tokens, we apply suppression within the neutral zone to reduce the influence of tokens lacking update tendency. This aligns with our analysis, where ratios far from 1 typically represent relatively important tokens, and we assign them relatively larger advantages:
\begin{equation}
\hat{A}_{i,t} = \begin{cases}
A_{i,t} \cdot (1 + \tilde{h}_{i,t}) & \text{if } C(\tilde{h}_{i,t}, r_{i,t}) \\
A_{i,t} & \text{otherwise}
\end{cases}
\end{equation}
\begin{equation}
C(\tilde{h}_{i,t}, r_{i,t}) = \begin{cases}
r_{i,t} \notin [\gamma_L, \gamma_U] & \text{if } \tilde{h}_t > 0  \\
r_{i,t} \in [\gamma_L, \gamma_U] & \text{if } \tilde{h}_t \leq 0
\end{cases}
\end{equation}

\subsection{Asymmetric Adaptive Clipping}
Unlike Archer ~\citep{wang2025stabilizingknowledgepromotingreasoning} that uniformly adjusts clipping bounds, we apply asymmetric adjustments respecting natural update tendencies of each token. By lowering the left boundary for low-entropy tokens, we allow these tokens to continue participating in the update process and are more likely to decrease substantially. By raising the right boundary for high-entropy tokens, we encourage exploration at critical decision points.
\begin{equation}
\epsilon_L({i,t}) = \begin{cases}
\epsilon_L^{\text{base}}(1- \tilde{h}_{i,t}) & \text{if } \tilde{h}_{i,t} \leq 0 \\
\epsilon_L^{\text{base}} & \text{if } \tilde{h}_{i,t} > 0
\end{cases}
\end{equation}
\begin{equation}
\epsilon_R({i,t}) = \begin{cases}
\epsilon_R^{\text{base}} & \text{if } \tilde{h}_{i,t} \leq 0 \\
\epsilon_R^{\text{base}}(1+ \tilde{h}_{i,t}) & \text{if } \tilde{h}_{i,t} > 0
\end{cases}
\end{equation}
where $\epsilon_L^{\text{base}}, \epsilon_R^{\text{base}}$ are the base clipping bounds. 

\begin{table*}[!t]
    \small
    \setlength{\extrarowheight}{3pt}
    \setlength{\tabcolsep}{7pt}
    \renewcommand{\arraystretch}{0.75}
    \centering
        \caption{Comparison between \emph{HAPO} and existing methods on Code and Logic benchmarks.}
        \label{tab: code_results}
    \resizebox{1.0\textwidth}{!}{%
    \begin{tabular}{@{}cc|ccccc|ccc@{}}
    \toprule
    \multirow{2}{*}{\textbf{Model}} & \multirow{2}{*}{\textbf{Method}} & \multicolumn{5}{c|}{\textbf{Logic RL}} & \multicolumn{3}{c}{\textbf{LiveCodeBench}} \\
    \cmidrule(lr){3-7} \cmidrule(l){8-10}
    & & \textbf{Avg} & 4ppl & 5ppl & 6ppl & 7ppl & \textbf{Avg} & V5(24/8/1-25/2/1) & V6(25/2/1-25/5/1) \\
    \midrule
    
     \multirow{5}{*}{\begin{tabular}[c]{@{}l@{}}Qwen2.5-7B-Instruct-1M\end{tabular}}    
     & Vanilla DAPO & 87.13 & 92.23 & 90.16 & 85.72 & 80.42 & 29.08 & 28.22 & 29.94\\
     
     & DAPO w/ Forking Tokens  & 90.15 & 93.44 & 92.13 & 90.26 & 84.75 & 31.46 & 29.72 & 33.19\\
     
     & Archer & 87.68 & 94.12 & 89.75 & 85.28 & 81.55 & 29.75 & 28.63 & 30.86\\
    
     & EDGE-GRPO & 88.92 & 92.35 & 91.13 & 89.58 & 82.63 & 30.40 & 29.27 & 31.53\\
     
     & HAPO(ours) & \textbf{92.69} & \textbf{95.41} & \textbf{93.34} & \textbf{93.26} & \textbf{88.74} & \textbf{34.02} & \textbf{32.55} & \textbf{35.48}\\
     
    \bottomrule
    \end{tabular}
    }
\end{table*}
\begin{table*}[!t]
    \small
    \setlength{\extrarowheight}{3pt}
    \setlength{\tabcolsep}{8pt}
    \renewcommand{\arraystretch}{0.75}
    \centering
        \caption{Component ablation of HAPO. We denote Adaptive Temperature Sampling, Token-Level Group Average Advantage Estimation, Differential Advantage Redistribution, and Asymmetric Adaptive Clipping as A, B, C, and D respectively for convenience.}
        \label{tab: component_abl}
    \resizebox{1.0\textwidth}{!}{%
    \begin{tabular}{@{}ccccccccccc@{}}
    \toprule
    \textbf{A} & \textbf{B} & \textbf{C} & \textbf{D} & \textbf{Avg (1560)} & AIME24 (30) & AIME25 (30) & AMC (83) & Math (500) &  OlympiadBench(675) & Minerva (272) \\
    \cmidrule(r){1-4} \cmidrule(l){5-11}

    & & & & 46.97 & 37.24 & 20.24 & 81.99 & 76.72 & 34.35 & 31.28 \\
        
     \ding{51}& & & & 48.85 & 39.77 & 23.01 & 84.20 & 76.88 & \textbf{37.34} & 31.92 \\
     
     & \ding{51}& & & 48.56 & 39.04 & 23.77 & 83.59 & 77.19 & 35.02 & 32.76 \\
     
     & & \ding{51}& & 48.28 & 38.63 & 22.07 & 84.19 & 76.58 & 36.00 & 32.19 \\
     
     & & & \ding{51}& 48.02 & 38.38 & 20.73 & 82.45 & 78.56 & 36.49 & 31.51 \\

    \midrule

     & & \ding{51}& \ding{51}& 48.74 & 39.19 & 22.55 & 84.43 & 77.72 & 36.13 & 32.44 \\

    & \ding{51}& \ding{51}& \ding{51}& 49.42 & 40.17 & 23.62 & 84.41 & \textbf{79.10} & 35.96 & 33.28 \\

    \ding{51}& \ding{51}& \ding{51}& \ding{51}& \textbf{50.04} & \textbf{41.31} & \textbf{24.34} & \textbf{85.47} & 78.45 & 36.94 & \textbf{33.73} \\
     
    \bottomrule
    \end{tabular}
    }
    \label{tab:componet_ablation}
\end{table*}

These continuous formulations unify our heterogeneous strategy, with all modulations smoothly varying as functions of normalized entropy $\tilde{h}$, ensuring a concise implementation. As entropy is already computed during sampling, HAPO introduces virtually no additional computational overhead.
We present the pseudocode of HAPO in Appendix ~\ref{app: method_alg}.

\section{Experiments}
\label{Experiments}
\textbf{Experimental Setup.} \label{subsec: experimental setup}
We incorporate our token-level strategy into DAPO~\citep{yu2025dapoopensourcellmreinforcement} within verl~\citep{verl} and vLLM~\citep{kwon2023efficientmemorymanagementlarge}. We maintain DAPO's settings: clip-higher of $\epsilon_{\text{high}}=0.28$ and $\epsilon_{\text{low}}=0.2$; overlong reward shaping with 10240-token maximum length and 4096-token cache. We use a training batch size of 512 and a mini-batch size of 32. Training proceeds with a learning rate of $10^{-6}$.

For HAPO's token-level strategies, we follow ~\citep{wang20258020rulehighentropyminority} and set the quantile $\rho$ to 80\%, encouraging exploration to the top 20\% highest-entropy tokens. For Adaptive Temperature Sampling, we compute $\hat\rho_{\log(H)}$ and $\hat\sigma_{\log(H)}$ based on all tokens' entropy in the previous step. We set $T_{\text{base}} = 1.0$ and $\tau = 0.1$. For Differential Advantage Redistribution, we determine the neutral zone using the clipping ratios that correspond to each token's actual dynamics, setting it to $[1-\frac{\epsilon_{L}}{2}, 1+\frac{\epsilon_{R}}{2}]$. For Asymmetric Adaptive Clipping, we set $\epsilon_L^{\text{base}} = 0.2$ and $\epsilon_R^{\text{base}} = 0.28$. We adopt a fixed hyperparameter setting across all models to explicitly showcase the generalizability of our method. We also provide hyperparameter ablations in Appendix~\ref{sec:additional_ablations}.

We perform experiments across Qwen2.5-Math-1.5B,Qwen2.5-Math-7B~\citep{yang2024qwen25mathtechnicalreportmathematical} base models and Qwen3-8B,Qwen3-14B~\citep{yang2025qwen3technicalreport} base models, using DAPO-Math-17K~\citep{yu2025dapoopensourcellmreinforcement} as the training dataset. We trained all models on 4 nodes with 32 A100 GPUs.

\textbf{Evaluation.} 
We evaluate all the models on mathematical benchmarks: AIME24, AIME25, AMC23, Minerva~\citep{lewkowycz2022solvingquantitativereasoningproblems}, MATH500~\citep{math500}, and OlympiadBench~\citep{olympiadbench}. For each question, we generate 8 responses under a decoding temperature $T=0.5$, and report the average accuracy.

\textbf{Main Results.} 
We compare HAPO against vanilla GRPO, vanilla DAPO, DAPO with Forking Tokens, Archer, Entropy Env and EDGE-GRPO in Table \ref{tab: qwen_math_results}.  On Qwen2.5-Math-7B, HAPO surpasses DAPO w/ Forking Tokens by 2.86 and 2.44 points on AIME'24 and AIME'25. When compared to entropy-based approaches, HAPO outperforms Archer by 2.80-4.41 points and Entropy Adv by 2.37-4.25 points, highlighting the effectiveness of our fine-grained heterogeneous treatment. Overall, HAPO outperforms these methods across models of different scales, demonstrating strong scalability. We visualize training dynamics in Figures ~\ref{fig: aime24_results},~\ref{fig: aime25_results},  ~\ref{fig: response_length}, comparing HAPO and DAPO throughout the training process. Our method maintains longer response lengths and higher accuracy, indicating that HAPO preserves model exploration capabilities while achieving better task performance.

\section{Ablations}
\label{ablations}
\textbf{Results on Models Other Than Qwen.}
We employ same experiments on LLaMA models. Results are presented in Table \ref{tab: llama_results}. HAPO demonstrates more pronounced improvements on weaker models, suggesting that our fine-grained heterogeneous treatment effectively helps models capture critical information, further validating HAPO's scalability. In Figure \ref{fig: llama_training_dynamics}, we show training dynamics of LLaMA3.2-3B. HAPO exhibits higher entropy and longer responses, enabling the model to maintain continuous exploration.

\textbf{Generalization to Other Domains.}
We validate HAPO's generalizability in code and logic. We use code data from Archer~\citep{wang2025stabilizingknowledgepromotingreasoning} and logic data from Logic-RL~\citep{xie2025logicrlunleashingllmreasoning}. We employ Qwen2.5-7B-Instruct-1M due to its general capabilities and the fact that it has not been pre-trained on these datasets. Evaluation on LiveCodeBench~\citep{jain2024livecodebenchholisticcontaminationfree} and Logic in Table \ref{tab: code_results} shows that HAPO consistently outperforms DAPO w/ Forking Tokens. This demonstrates that token heterogeneity exists across different domains, and HAPO's approach naturally generalizes to diverse areas, confirming  HAPO's broad applicability.

\textbf{Component-wise Contributions.}
We examine each component's contribution in Table \ref{tab: component_abl}. All components contribute to HAPO's performance, with Adaptive Temperature Sampling proving particularly crucial as it governs token distribution and entropy. Token-Level Group Average Advantage Estimation significantly impacts advantage redistribution and clipping, corroborating our analysis. In fact, all components form an organic unity: Adaptive Temperature Sampling ensures sufficient high-entropy critical tokens, Differential Advantage Redistribution optimizes them with larger updates, and Asymmetric Adaptive Clipping  encourages exploration. This synergy enables effective leverage of token heterogeneity throughout optimization. 


\section{Conclusion}
We introduce Heterogeneous Adaptive Policy Optimization (HAPO), which leverages entropy to drive the entire optimization process and implements fine-grained token-level optimization. Experiments demonstrate consistent improvements over DAPO. HAPO establishes that effective RLHF requires adapting to token heterogeneity, opening new ways for developing more capable reasoning models.

\section*{Limitations}
Despite HAPO's strong empirical performance, we acknowledge several limitations that provide opportunities for future refinement:

\textbf{Entropy as Primary Heterogeneity Signal.} While our entropy-based approach effectively identifies token heterogeneity, it may not capture all aspects of token importance. Tokens with moderate entropy could still be semantically crucial, particularly in structured or deterministic outputs. Future work could explore complementary signals like gradient magnitudes or attention weights.

\textbf{Static Entropy Thresholds.} HAPO uses a fixed 80th percentile to distinguish high/low entropy tokens throughout training, though entropy distributions evolve as training progresses. Dynamic threshold adaptation could potentially improve performance but would add complexity.

These limitations suggest directions for future refinement.

\section*{Acknowledgments}
This work is supported by National Natural Science Foundation of China (92470121, 62402016), Fundamental and Interdisciplinary Disciplines Breakthrough Plan of the Ministry of Education of China (JYB2025XDXM113), National Key R\&D Program of China (2024YFA1014003), Zhongguancun Academy (C20250204, C20250602),  Beijing Major Science and Technology Project (Z251100008125043, Z251100008425023), and High-performance Computing Platform of Peking University.

\bibliography{custom}

\newpage
\clearpage
\appendix

\begin{center}
    \Large{\textbf{Appendix}}
\end{center}
\nolinenumbers

\section{Related Work}
\subsection{Reinforcement Learning from Human Feedback}
The evolution of reinforcement learning in language models represents a trajectory from basic preference alignment toward sophisticated reasoning capabilities. Foundational work in constrained optimization emerged through TRPO ~\citep{schulman2017trustregionpolicyoptimization} and PPO ~\citep{schulman2017proximalpolicyoptimizationalgorithms}, establishing principles for stable policy updates. Subsequent algorithmic innovations eliminated computational bottlenecks—GRPO ~\citep{shao2024deepseekmathpushinglimitsmathematical} and REINFORCE++ ~\citep{hu2025reinforceefficientrlhfalgorithm} removed value network dependencies via group-based advantage computation. The offline optimization paradigm gained prominence through DPO ~\citep{rafailov2024directpreferenceoptimizationlanguage}, KTO ~\citep{ethayarajh2024ktomodelalignmentprospect}, and SimPO ~\citep{meng2024simposimplepreferenceoptimization}, which circumvent reward model training entirely.

A fundamental transformation occurred with the advent of reasoning-centric models. OpenAI's o1 ~\citep{o1} pioneered effective multi-step reasoning through reinforcement learning at scale, catalyzing widespread development. DeepSeek-R1 ~\citep{deepseekai2025deepseekr1incentivizingreasoningcapability} demonstrated reasoning emergence without supervised fine-tuning, while Claude3.5 ~\citep{claude35sonnet}, Qwen3 ~\citep{yang2025qwen3technicalreport}, and Seed-1.5-Thinking ~\citep{seed2025seed15thinkingadvancingsuperbreasoning} expanded the reasoning frontier~\citep{liu2025fusion,liu2025synthvlm,lin2026mmfinereason,liu2026chartverse,zhang-etal-2025-ga,zhang2026semanticawarelogicalreasoningsemiotic,zhang2026logicalphasetransitionsunderstanding,zhang2026couplingmacrodynamicsmicro}. Beyond these state-of-the-art models, various works explored complementary optimization strategies, including SimpleRLZoo ~\citep{zeng2025simplerlzooinvestigatingtamingzero} and Open-Reasoner-Zero, which investigate alternative training paradigms. Meanwhile, algorithmic refinements continue to enhance existing methods—DAPO ~\citep{yu2025dapoopensourcellmreinforcement} introduced token-mean averaging and asymmetric clipping for extended sequences, while VAPO ~\citep{yue2025vapoefficientreliablereinforcement} developed variance-aware optimization strategies to improve training stability and convergence.

\subsection{Entropy-Based Optimization in LLMs}
Entropy has emerged as a fundamental signal for quantifying model uncertainty and identifying token heterogeneity in reasoning tasks. Pioneering works have established the theoretical and empirical basis for entropy-aware learning~\citep{zeng2025simplerlzooinvestigatingtamingzero,liu2025understandingr1zeroliketrainingcritical,cai2025opendataarena,cui2025entropymechanismreinforcementlearning,lin2025se,li2025curriculum,ReTrack,HABIT}. analyzed how entropy dynamics correlate with reasoning capabilities, while some works ~\citep{wang20258020rulehighentropyminority} proposed the "80/20 rule," demonstrating that a minority of high-entropy "forking tokens" effectively control reasoning diversity. Other studies~\citep{agarwal2025unreasonableeffectivenessentropyminimization,gao2025oneshotentropyminimization,cai2025opendataarena,zhu2026factors,nie2026attnpo,wu2026steppotentialadvantageestimation,zhou2026look,ZHANG2026112674,REFINE,HINT} have shown that entropy minimization alone can serve as a powerful unsupervised signal.

Building upon these identification strategies, recent research has actively explored incorporating entropy into optimization mechanisms, typically following two paradigms:

\textbf{Differentiated Clipping and Categorization.} One line of work focuses on relaxing constraints for critical tokens. DAPO with forking tokens~\citep{wang20258020rulehighentropyminority} and Archer~\citep{wang2025stabilizingknowledgepromotingreasoning,MELT,STABLE} employ entropy to categorize tokens into high and low entropy groups, adjusting clipping boundaries to encourage exploration. However, these methods generally function as discrete filters, relying on rigid thresholds that create artificial discontinuities in the optimization landscape.

\begin{figure*}[!t]
    \begin{subfigure}{0.24\textwidth}
        \centering
        \includegraphics[height=2.2cm,width=3.2cm]{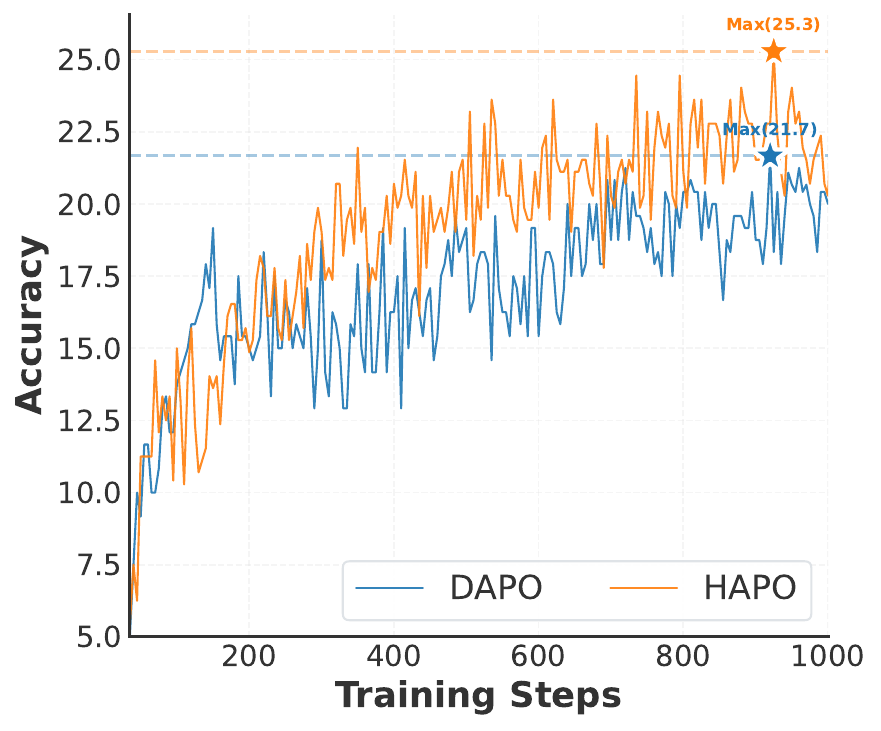}
        \caption{Qwen2.5-Math-1.5B}
    \end{subfigure}
    \hfill
    \begin{subfigure}{0.24\textwidth}
        \centering
        \includegraphics[height=2.2cm,width=3.2cm]{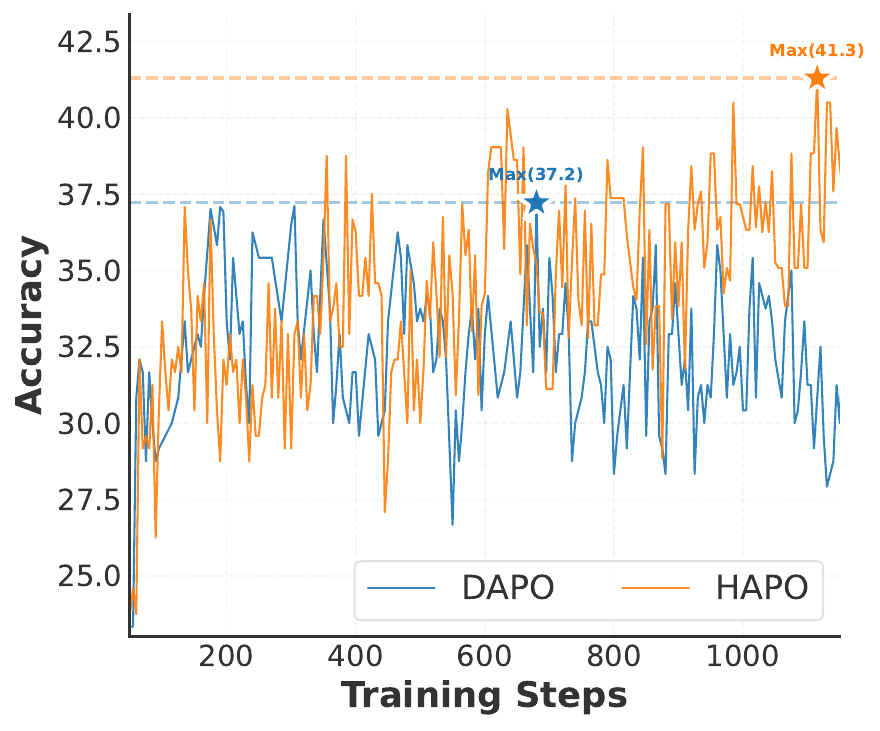}
        \caption{Qwen2.5-Math-7B}
    \end{subfigure}
        \begin{subfigure}{0.24\textwidth}
        \centering
        \includegraphics[height=2.2cm,width=3.2cm]{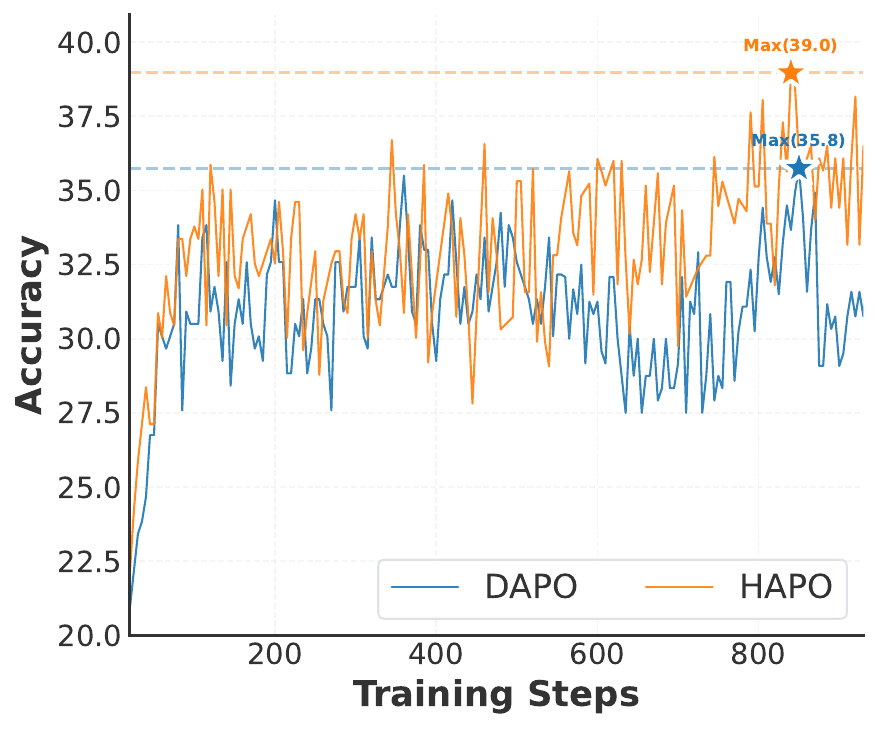}
        \caption{Qwen3-8B}
    \end{subfigure}
    \hfill
    \begin{subfigure}{0.24\textwidth}
        \centering
        \includegraphics[height=2.2cm,width=3.2cm]{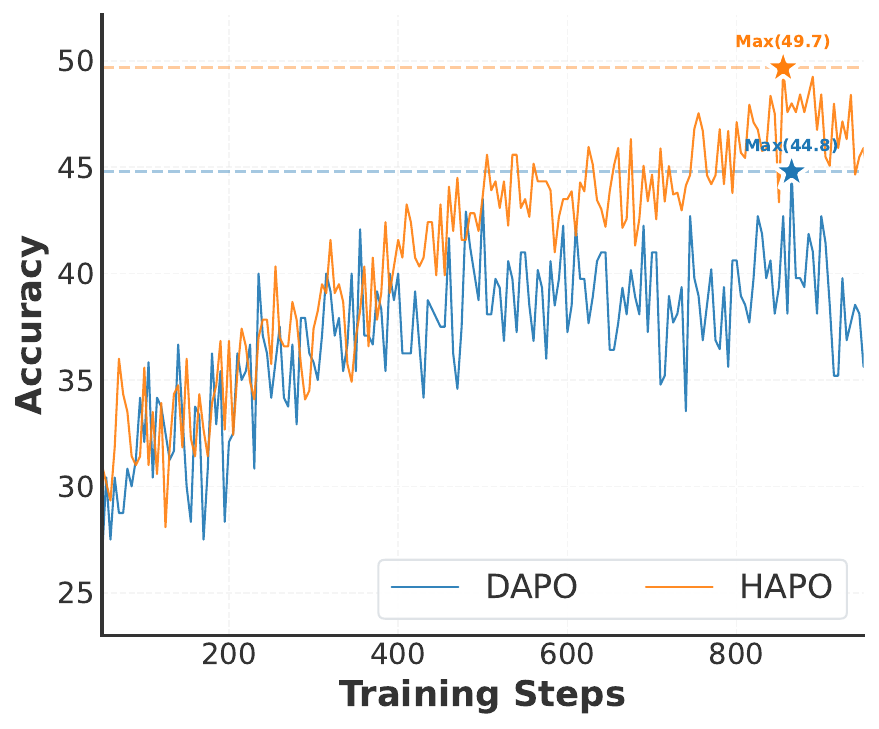}
        \caption{Qwen3-14B}
    \end{subfigure}
    \caption{Training dynamics of AIME24 results for HAPO and DAPO.}
    \label{fig: aime24_results}
\end{figure*}
\begin{figure*}[!t]
    \begin{subfigure}{0.24\textwidth}
        \centering
        \includegraphics[height=2.2cm,width=3.2cm]{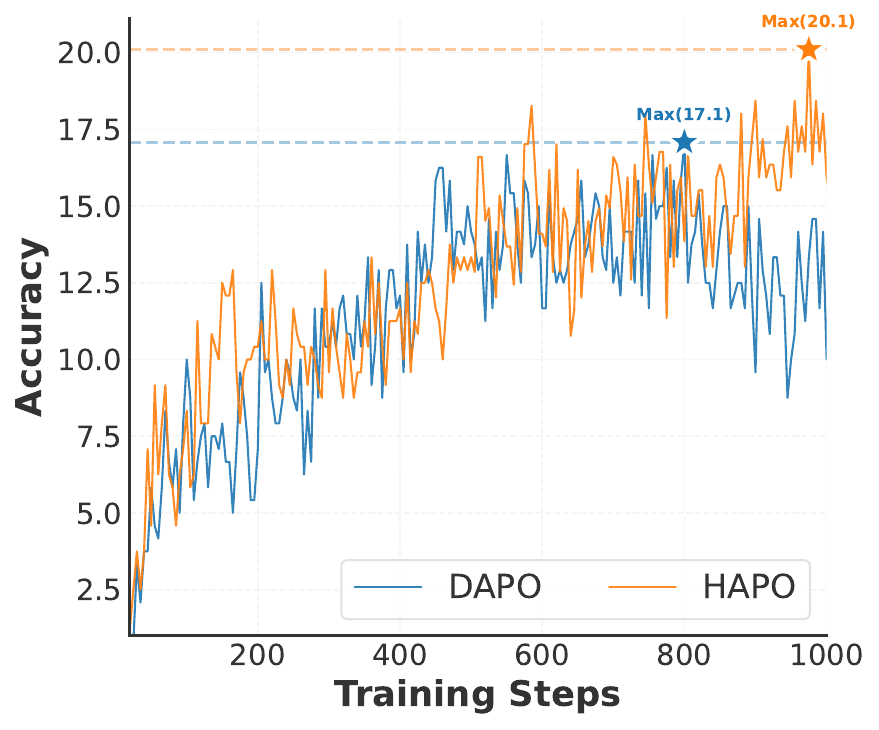}
        \caption{Qwen2.5-Math-1.5B}
    \end{subfigure}
    \hfill
    \begin{subfigure}{0.24\textwidth}
        \centering
        \includegraphics[height=2.2cm,width=3.2cm]{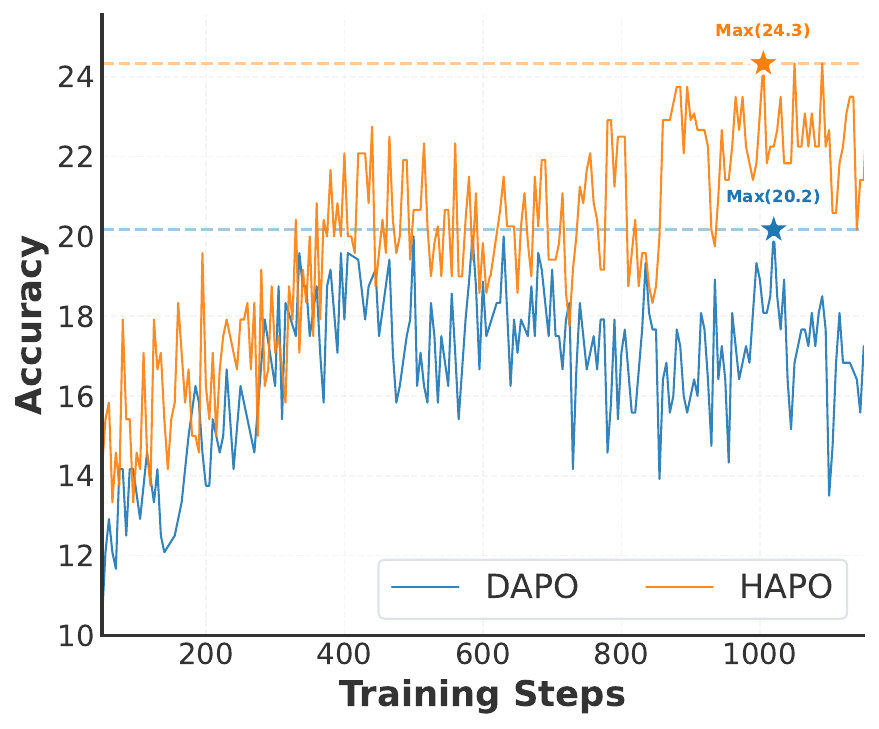}
        \caption{Qwen2.5-Math-7B}
    \end{subfigure}
        \begin{subfigure}{0.24\textwidth}
        \centering
        \includegraphics[height=2.2cm,width=3.2cm]{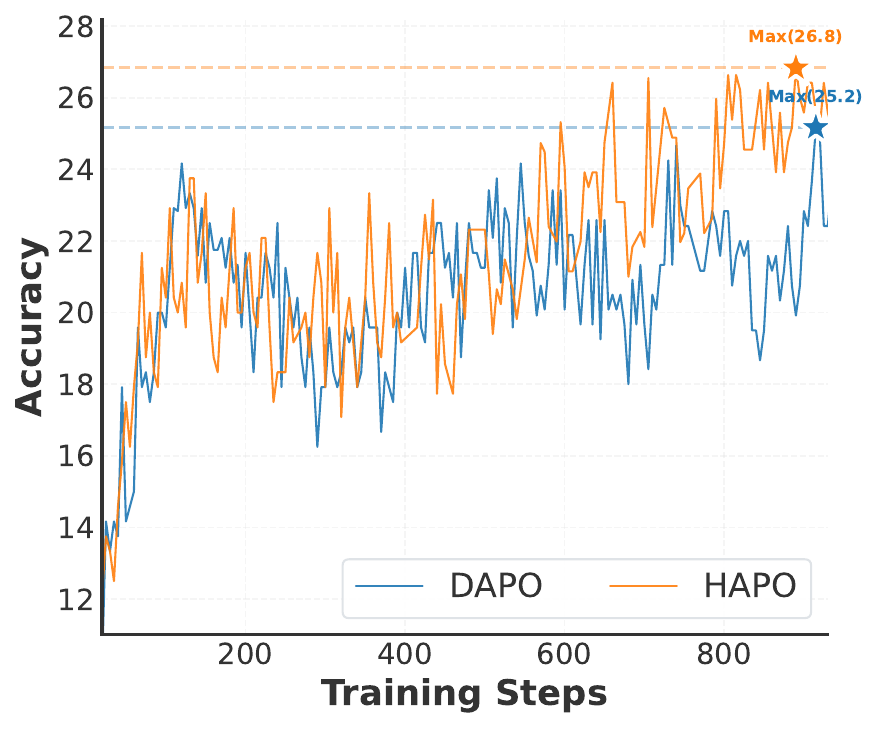}
        \caption{Qwen3-8B}
    \end{subfigure}
    \hfill
    \begin{subfigure}{0.24\textwidth}
        \centering
        \includegraphics[height=2.2cm,width=3.2cm]{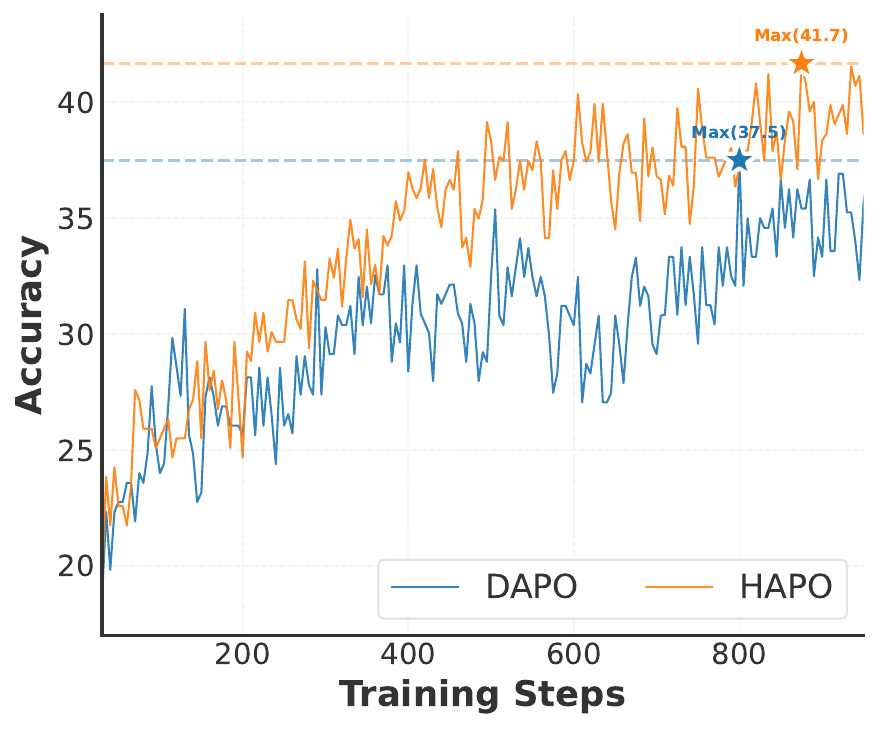}
        \caption{Qwen3-14B}
    \end{subfigure}
    \caption{Training dynamics of AIME25 results for HAPO and DAPO.}
    \label{fig: aime25_results}
\end{figure*}
\begin{figure*}[!t]
    \begin{subfigure}{0.24\textwidth}
        \centering
        \includegraphics[height=2.2cm,width=3.2cm]{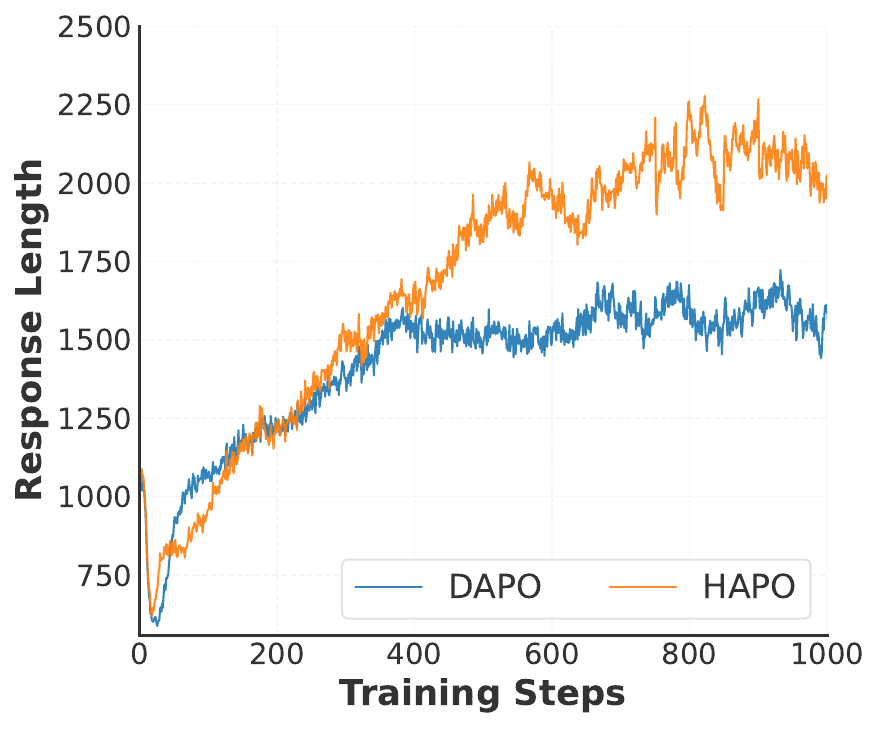}
        \caption{Qwen2.5-Math-1.5B}
    \end{subfigure}
    \hfill
    \begin{subfigure}{0.24\textwidth}
        \centering
        \includegraphics[height=2.2cm,width=3.2cm]{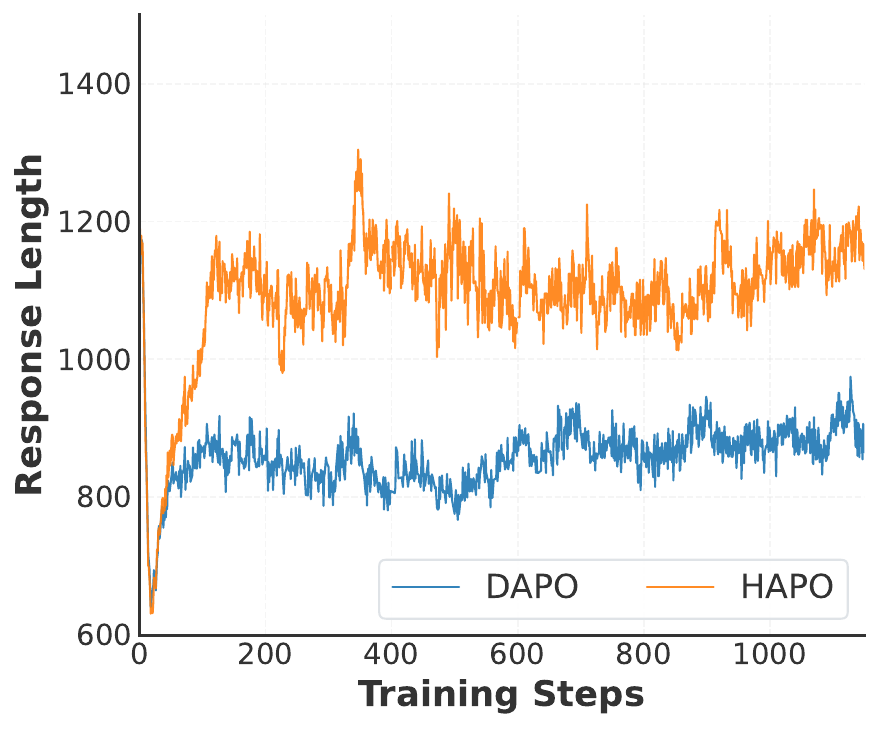}
        \caption{Qwen2.5-Math-7B}
    \end{subfigure}
        \begin{subfigure}{0.24\textwidth}
        \centering
        \includegraphics[height=2.2cm,width=3.2cm]{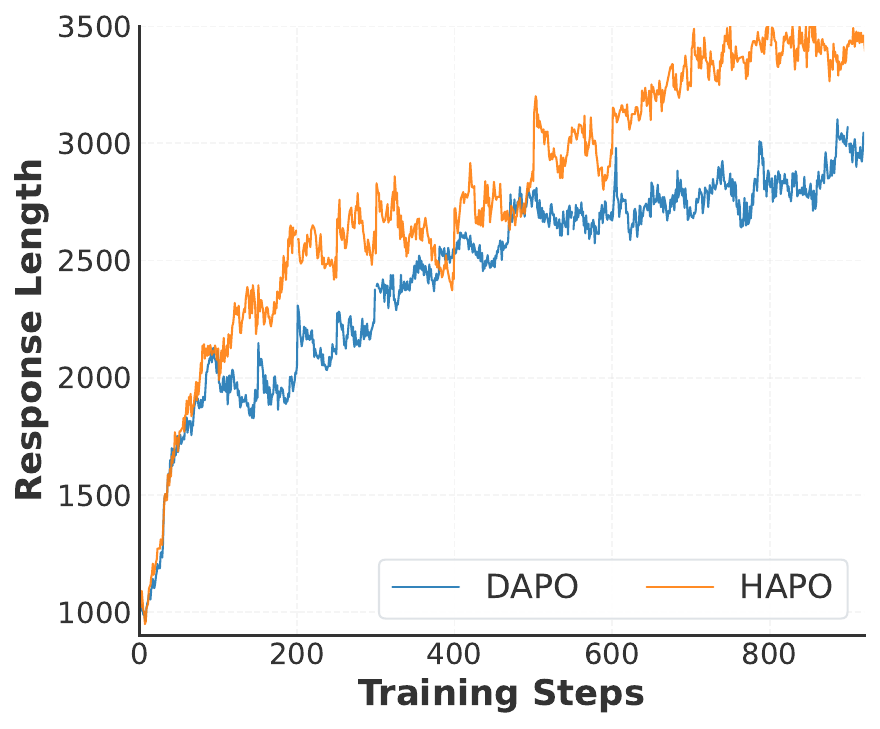}
        \caption{Qwen3-8B}
    \end{subfigure}
    \hfill
    \begin{subfigure}{0.24\textwidth}
        \centering
        \includegraphics[height=2.2cm,width=3.2cm]{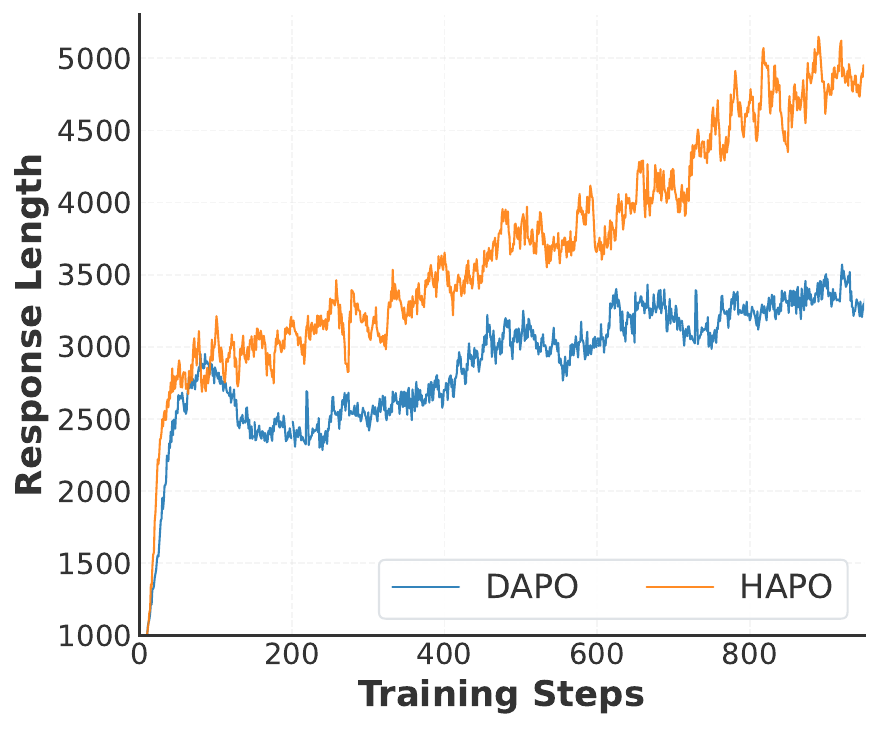}
        \caption{Qwen3-14B}
    \end{subfigure}
    \caption{Training dynamics of response length for HAPO and DAPO.}
    \label{fig: response_length}
\end{figure*}

\begin{figure*}[!t]
    \begin{subfigure}{0.24\textwidth}
        \centering
        \includegraphics[height=2.3cm,width=3.2cm]{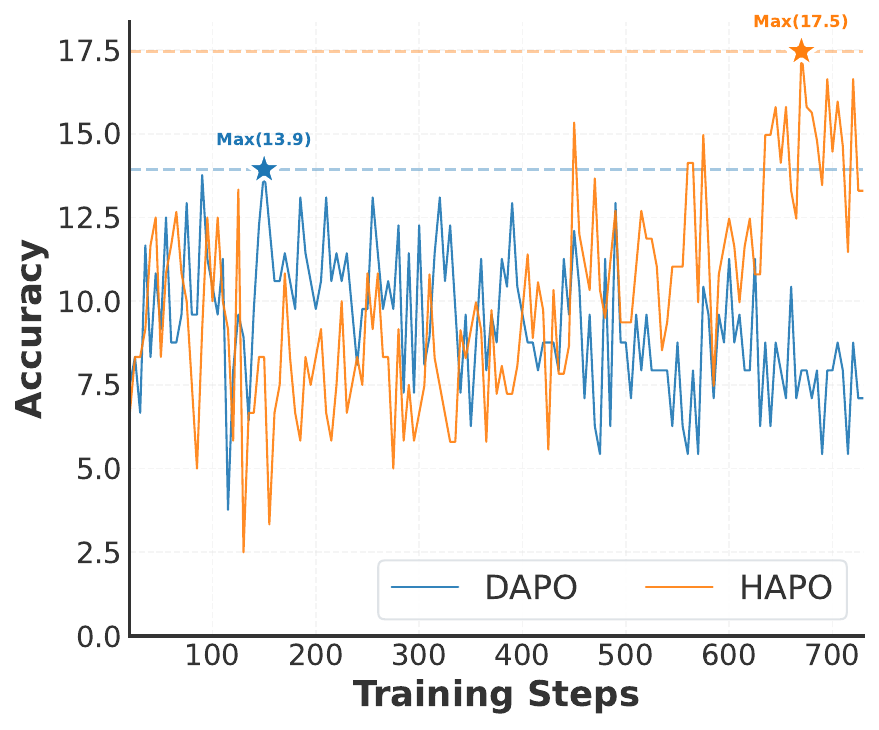}
        \caption{AIME24}
    \end{subfigure}
    \hfill
    \begin{subfigure}{0.24\textwidth}
        \centering
        \includegraphics[height=2.3cm,width=3.2cm]{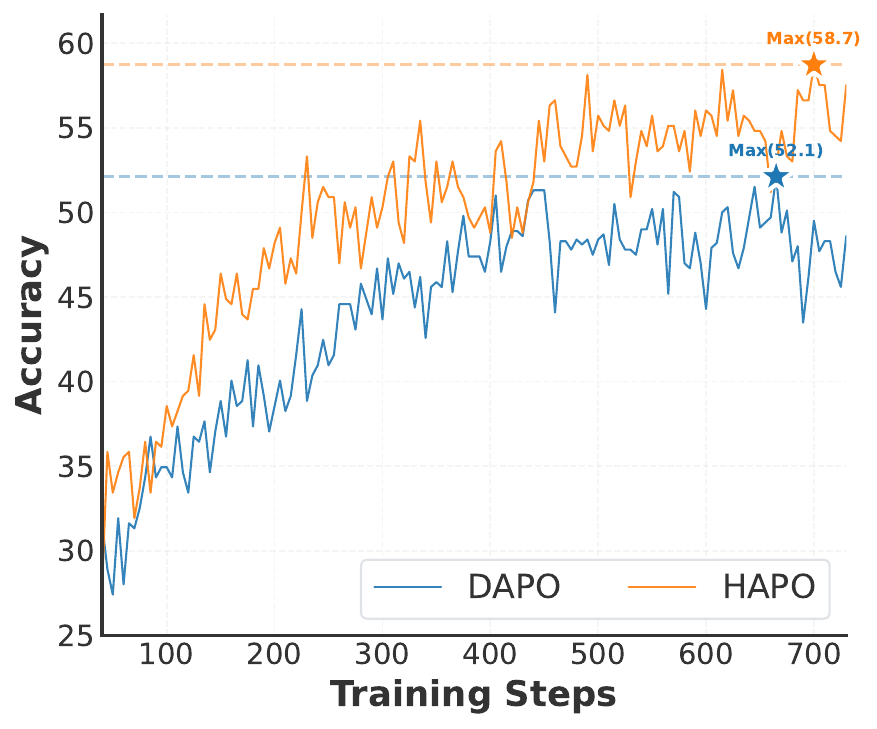}
        \caption{AMC}
    \end{subfigure}
        \begin{subfigure}{0.24\textwidth}
        \centering
        \includegraphics[height=2.3cm,width=3.2cm]{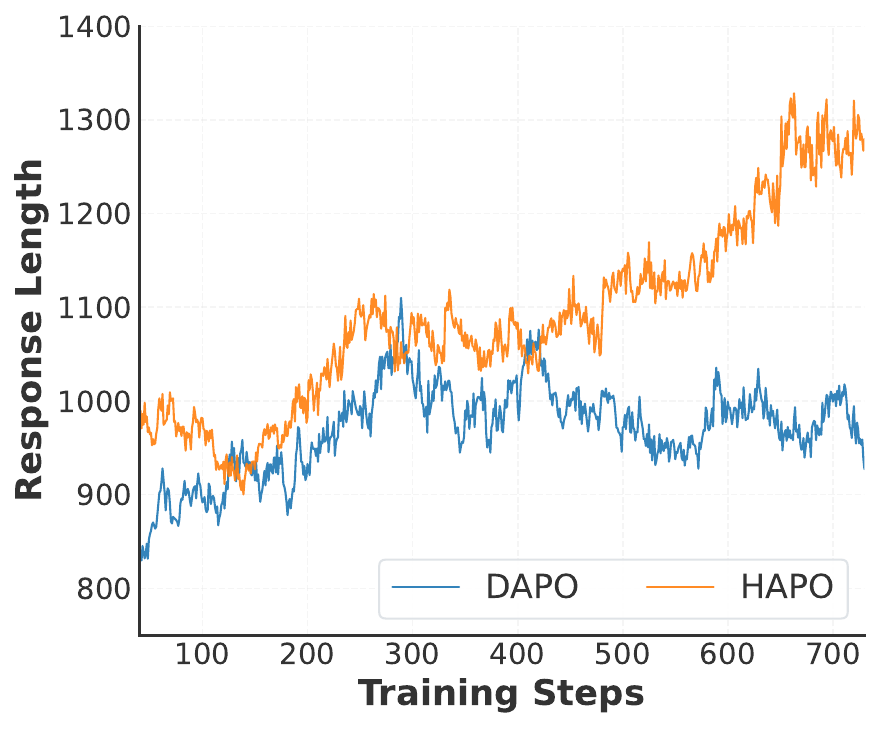}
        \caption{Response Length}
    \end{subfigure}
    \hfill
    \begin{subfigure}{0.24\textwidth}
        \centering
    \includegraphics[height=2.3cm,width=3.2cm]{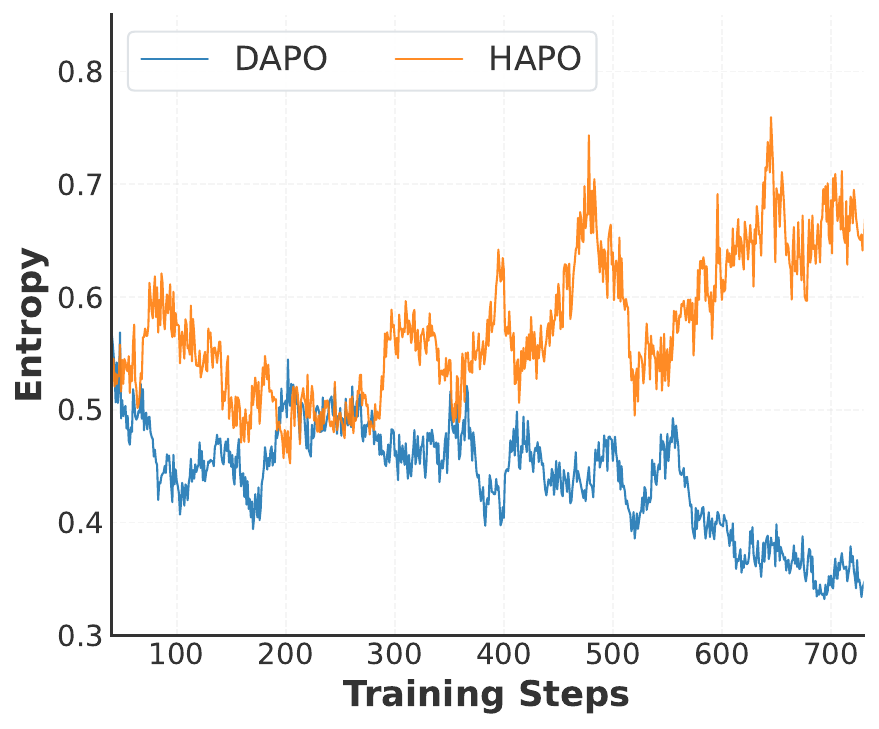}
        \caption{Entropy}
    \end{subfigure}
    \caption{Training dynamics of HAPO  and DAPO on LLaMA3.2-3B-Instruct.}
    \label{fig: llama_training_dynamics}
\end{figure*}

\textbf{Advantage Modulation and Regularization.} Another direction involves reshaping learning signals directly. Entropy Adv~\citep{cheng2025reasoningexplorationentropyperspective,} and EDGE-GRPO~\citep{zhang2025edgegrpoentropydrivengrpoguided,OFFSET,INTENT} propose modulating advantages based on entropy, assigning higher weights or bonuses to confident or high-entropy tokens. While effective, these approaches typically treat entropy as an additive bonus or an auxiliary regulator. By superimposing entropy onto existing rewards rather than integrating it into the core functional dependencies of optimization, they often function as post-hoc regularizers.

\section{Training Dynamics of HAPO}
Figures ~\ref{fig: aime24_results}, ~\ref{fig: aime25_results}, and ~\ref{fig: response_length} illustrate the training dynamics of HAPO versus DAPO across Qwen2.5 and Qwen 3 models, focusing on response length, entropy, and accuracy. Our observations reveal that HAPO sustains longer reasoning chains and higher accuracy levels, suggesting that it balances extensive exploration with precise execution. Additionally, comparative analysis on the LLaMA series in Figure ~\ref{fig: llama_training_dynamics} shows that HAPO exhibits robust training dynamics, further confirming its generalizability.

\section{HAPO Pseudocode}
\label{app: method_alg}
Figure~\ref{fig:hapo_code} provides the complete pseudocode of HAPO based on the verl and vLLM implementation. The algorithm details the step-by-step process of: (1) computing entropy statistics for continuous regulation, (2) applying adaptive temperature during sampling, (3) performing token-level group average advantage estimation, (4) implementing differential advantage redistribution based on importance ratios and entropy values, and (5) executing asymmetric adaptive clipping for policy updates. 

The pseudocode highlights HAPO's key innovation: using normalized entropy $\tilde{h}_{i,t} \in [-1, 1]$ as a continuous regulator that smoothly modulates all components without introducing significant computational overhead. The entropy computation required for sampling is reused throughout the pipeline, ensuring efficient implementation. The algorithm maintains simplicity while achieving fine-grained heterogeneous treatment, making it applicable to existing RL frameworks.

\begin{figure*}[!t]
    \centering
    \includegraphics[width=16cm]{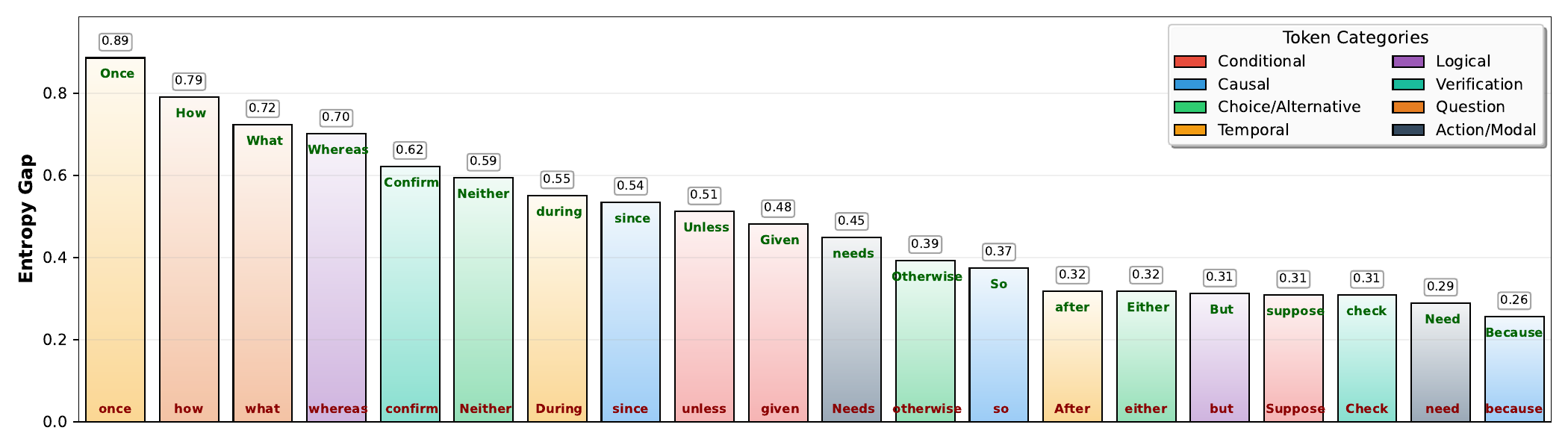}
    \caption{Dual-Entropy Token Frequency-Entropy Landscape}
    \label{fig:max_20_entropy}
\end{figure*}

\begin{table*}[t]
    \small
    \setlength{\extrarowheight}{3pt}
    \setlength{\tabcolsep}{6pt}
    \renewcommand{\arraystretch}{0.9}
    \centering
    \caption{Ablation study on entropy quantile threshold $\rho$ using Qwen2.5-Math-7B. The 80th percentile achieves optimal balance.}
    \label{tab: entropy_quantile_abl}
    \scalebox{0.93}{
    \begin{tabular}{@{}c|ccccccc@{}}
    \toprule
    \textbf{Quantile} & \textbf{Avg (1560)} & AIME24 (30) & AIME25 (30) & AMC (83) & Math (500) &  OlympiadBench(675) & Minerva (272) \\
    \midrule
    90\% & 48.96 & 39.44 & 22.73 & 84.11 & 77.74 & 36.61 & 33.14 \\
    \textbf{80\%} & \textbf{50.04} & \textbf{41.31} & \textbf{24.34} & \textbf{85.47} & 78.45 & 36.94 & \textbf{33.73} \\
    70\% & 49.74 & 40.78 & 24.11 & 84.96 & \textbf{78.83} & \textbf{37.05} & 32.71 \\
    50\% & 48.45 & 38.92 & 23.67 & 82.41 & 76.98 & 36.42 & 32.35 \\
    30\% & 47.10 & 37.05 & 23.12 & 81.58 & 75.21 & 34.39 & 31.26 \\
    \bottomrule
    \end{tabular}
    }
\end{table*}


\section{Additional Hyperparameter Ablations}
\label{sec:additional_ablations}

We conduct extensive ablation studies to validate our hyperparameter choices and design decisions. All experiments in this section are performed on Qwen2.5-Math-7B following the same training protocol described in Section \ref{subsec: experimental setup}. 

\textbf{Entropy Quantile Threshold.} We investigate the impact of the entropy quantile threshold $\rho$, which determines the boundary between high and low entropy tokens. Table \ref{tab: entropy_quantile_abl} presents results across different quantile values on Qwen2.5-Math-7B.

The results confirm that the 80th percentile provides optimal performance. The 70th percentile achieves comparable results, suggesting robustness around this range. Setting $\rho$ to 90\% degrades performance as this restrictive threshold identifies too few high-entropy tokens (only top 10\%), limiting exploration at critical points. Conversely, lower thresholds (50\% and 30\%) progressively worsen performance by classifying too many tokens as high-entropy, diluting the focused amplification effect. The 30\% threshold particularly suffers, treating most tokens as high-entropy and applying unfocused modifications. This validates the 80th percentile as balancing critical token identification with sufficient coverage.

\textbf{Temperature Adjustment Factor.} The temperature adjustment factor $\tau$ controls the magnitude of dynamic temperature modulation. We evaluate its impact across a range of values in Table \ref{tab: temperature_abl}.

Our chosen value of $\tau=0.1$ achieves the best overall performance. While $\tau=0.05$ produces competitive results, this smaller adjustment factor limits the temperature modulation range too narrowly, preventing the mechanism from fully leveraging adaptive temperature benefits. The complete absence of temperature adjustment ($\tau=0$) eliminates this exploration mechanism entirely. Conversely, larger values ($\tau \geq 0.3$ and $\tau \geq 0.5$) lead to excessive temperature modifications. Since high-entropy tokens inherently exhibit strong randomness, applying overly aggressive temperature increases to these already uncertain positions introduces extreme instability, disrupting coherent reasoning chains. This demonstrates the importance of balanced temperature modulation—sufficient to encourage exploration at critical points without transforming productive exploration into chaotic sampling.

\begin{table*}[!t]
    \small
    \setlength{\extrarowheight}{3pt}
    \setlength{\tabcolsep}{6pt}
    \renewcommand{\arraystretch}{0.9}
    \centering
    \caption{Ablation study on temperature adjustment factor $\tau$ using Qwen2.5-Math-7B. Moderate adjustment ($\tau=0.1$) yields best results.}
    \label{tab: temperature_abl}
    \scalebox{0.93}{
    \begin{tabular}{@{}c|ccccccc@{}}
    \toprule
    \textbf{$\tau$} & \textbf{Avg (1560)} & AIME24 (30) & AIME25 (30) & AMC (83) & Math (500) &  OlympiadBench(675) & Minerva (272) \\
    \midrule
    0.0 & 49.42 & 40.17 & 23.62 & 84.41 & \textbf{79.10} & 35.96 & 33.28 \\
    0.05 & 49.58 & 40.86 & 24.05 & 83.98 & 78.74 & 36.52 & 33.31 \\
    \textbf{0.1} & \textbf{50.04} & \textbf{41.31} & \textbf{24.34} & \textbf{85.47} & 78.45 & \textbf{36.94} & \textbf{33.73} \\
    0.3 & 49.11 & 40.67 & 23.61 & 83.13 & 77.58 & 36.89 & 32.36 \\
    0.5 & 48.12 & 38.92 & 22.45 & 83.27 & 76.83 & 35.14 & 32.08 \\
    \bottomrule
    \end{tabular}
    }
\end{table*}

\begin{table*}[!t]
    \small
    \setlength{\extrarowheight}{3pt}
    \setlength{\tabcolsep}{6pt}
    \renewcommand{\arraystretch}{0.9}
    \centering
    \caption{Ablation study on neutral zone configuration using Qwen2.5-Math-7B. Using half of the clipping bounds provides optimal control.}
    \label{tab: neutral_zone_abl}
    \scalebox{0.88}{
    \begin{tabular}{@{}cccccccc@{}}
    \toprule
    \textbf{Neutral Zone} & \textbf{Avg (1560)} & AIME24 (30) & AIME25 (30) & AMC (83) & Math (500) &  OlympiadBench(675) & Minerva (272) \\
    \midrule
    Full Clip Bounds & 49.57 & 40.77 & 24.13 & 84.86 & 78.23 & 36.48 & 32.93 \\
    \textbf{1/2 Clip Bounds} & \textbf{50.04} & 41.31 & \textbf{24.34} & \textbf{85.47} & 78.45 & \textbf{36.94} & \textbf{33.73} \\
    Zero Bounds & 49.71 & \textbf{41.42} & 23.51 & 85.23 & \textbf{78.62} & 36.17 & 33.32 \\
    No Zone & 49.37 & 40.83 & 23.67 & 84.58 & 77.94 & 36.56 & 32.61 \\
    \bottomrule
    \end{tabular}
    }
\end{table*}

\textbf{Neutral Zone Configuration.} The neutral zone $[\gamma_L, \gamma_U]$ determines when differential advantage redistribution is applied based on importance ratios. We examine various configurations based on the clipping bounds in Table \ref{tab: neutral_zone_abl}.

Setting the neutral zone to half of the clipping bounds achieves optimal performance. The No Zone configuration, which applies no differential advantage redistribution, performs worst—without entropy-based modulation, the model treats all tokens equally regardless of their importance. Full Clip Bounds, which only suppresses low-entropy token advantages without amplifying high-entropy ones, shows marginal improvement but fails to encourage exploration at critical decision points. Zero Bounds performs better by only amplifying high-entropy token advantages. While this promotes exploration, it neglects to suppress trivial low-entropy tokens, leading to inefficient updates. \textbf{1/2 Clip Bounds} configuration achieves the best results by combining both mechanisms: it suppresses low-entropy tokens within the neutral zone while amplifying high-entropy tokens whose importance ratios indicate meaningful updates. This balanced approach ensures that computational resources focus on genuinely important tokens while reducing noise from trivial updates, demonstrating the necessity of bidirectional entropy-aware modulation.

These comprehensive ablation studies on Qwen2.5-Math-7B confirm that our hyperparameter choices are well-calibrated and that each design decision contributes meaningfully to HAPO's strong performance. The continuous, entropy-based regulation framework proves essential for effectively leveraging token heterogeneity in reinforcement learning optimization.

\begin{figure*}[h]
\small
\begin{algorithmic}[1]
\Require Policy $\pi_\theta$, dataset $\mathcal{D}$
\Require Entropy quantile: $\rho$
\Require Temperature params: $T_{\text{base}}, \hat\rho_{\log(H)}, \hat\sigma_{\log(H)}, \tau$ \quad Clipping params: base bounds $\epsilon_L^{\text{base}}, \epsilon_R^{\text{base}}$
\Ensure Updated policy $\pi_{\theta'}$

\Comment{Sampling Rollout Sequences} 
\For{each token position $i,t$}
\Comment{Continuous Adaptive Temperature Sampling}
    \State Compute entropy: $H_{i,t} \gets -\sum_{v}\pi_{\theta}(v|s_{i,t})\log\pi_{\theta}(v|s_{i,t})$
    \State Adaptive temperature: $T_{i,t} \gets T_{\text{base}} \cdot \left(1 + \frac{\log(H_{i,t}) - \hat\rho_{\log(H)}}{\hat\sigma_{\log(H)}} \cdot \tau \right)$
\EndFor

    \State Assign rewards to tokens: $a_{i,t} \gets r_i$, $\mathcal{T} \gets \{(i,t)\}$
    \Comment{Token-Level Group Average Advantage Estimation}
    \State Compute token-level statistics: $\mu_{\text{tok}} \gets \frac{1}{|\mathcal{T}|}\sum_{(i,t)} a_{i,t}, \sigma_{\text{tok}} \gets \sqrt{\frac{1}{|\mathcal{T}|}\sum_{(i,t)} (a_{i,t} - \mu_{\text{tok}})^2}$
    \State Normalize advantages: $A_{i,t} \gets (a_{i,t} - \mu_{\text{tok}}) / \sigma_{\text{tok}}$
    \State $Q_{\rho}(\log(H)) \gets \text{Quantile}_{\rho}(\{\log(H_{i,t}) : (i,t) \in \mathcal{T}\})$
    \Comment{Compute global entropy statistics}
    \State $\sigma({\log(H)}) \gets \sqrt{\frac{1}{|\mathcal{T}|} \sum_{(i,t) \in \mathcal{T}} (\log(H_{i,t}) - Q_{\rho}(\log(H)))^2}$
\For{each mini batch $\mathcal{B}$}
\Comment{Training with Heterogeneous Treatment}
    \For{each token $i,t$ in $\mathcal{B}$}
        \State Compute log-entropies: $h_{i,t} = \frac{\log(H_{i,t}) - Q_{\rho}(\log(H))}{\sigma(\log(H))}$
        \State Asymmetric scaling: $\tilde{h}_{i,t} \gets \begin{cases} h_{i,t}/h_{\max} & \text{if } h_{i,t} > 0 \\ -h_{i,t}/|h_{\min}| & \text{if } h_{i,t} \leq 0 \end{cases}$
        
        \Comment{Continuous Asymmetric Adaptive Clipping - Compute First}
        
        \If{$\tilde{h}_{i,t} \leq 0$} 
            \State $\epsilon_L({{i,t}}) \gets \epsilon_L^{\text{base}} (1-\tilde{h}_{i,t}),\epsilon_R({{i,t}}) \gets \epsilon_R^{\text{base}}$
        \Else
            \State $\epsilon_L({{i,t}}) \gets \epsilon_L^{\text{base}},\epsilon_R({{i,t}}) \gets \epsilon_R^{\text{base}} (1+ \tilde{h}_{i,t})$
        \EndIf
    \State $\gamma_L \gets 1 - \frac{\epsilon_L({i,t})}{2}, \gamma_U \gets 1 + \frac{\epsilon_R({i,t})}{2}$
    \Comment{Neutral Zone based on Dynamic Clipping}
    
        \Comment{Continuous Differential Advantage Redistribution}
        \State Compute importance ratio: $r_{i,t} \gets \pi_\theta(a_{i,t}|s_{i,t}) / \pi_{\theta_{\text{old}}}(a_{i,t}|s_{i,t})$
        \If{$(\tilde{h}_{i,t} > 0 $ and $r_{i,t} \notin [\gamma_L, \gamma_U])$ or $(\tilde{h}_{i,t} \leq 0$ and $r_{i,t} \in [\gamma_L, \gamma_U]))$}
        \State $\hat{A}_{i,t} \gets A_{i,t} \cdot (1 + \tilde{h}_{i,t})$
        \Else
        \State $\hat{A}_{i,t} \gets A_{i,t}$
        \EndIf
    \EndFor

    \Comment{Compute HAPO loss}
    \State $\mathcal{L}^{\text{HAPO}}(\theta) = \left[ \frac{1}{\sum_{i=1}^G |o_i|} \sum_{i=1}^{G} \sum_{t=1}^{|o_i|} \min \left( r_{i,t}(\theta) \hat{A}_{i,t}, \text{clip}(r_{i,t}(\theta), 1-\epsilon_L(i,t), 1+\epsilon_R(i,t)) \hat{A}_{i,t} \right) \right]$
    \State Update parameters: $\theta' \gets \theta + \eta \cdot \nabla_{\theta}\mathcal{L}^{\text{HAPO}}(\theta)$
\EndFor

\State $\hat\rho_{\log(H)}\gets Q_{\rho}(\log(H_t)), 
\hat\sigma_{\log(H)} \gets \sigma(\log(H_t))$
\Comment{Update entropy statistics for next step}
\end{algorithmic}
\caption{HAPO algorithm overview. The method dynamically adjusts temperature, advantages, and clipping bounds based on normalized entropy $\tilde{h}_{i,t} \in [-1,1]$, implementing heterogeneous treatment for tokens throughout the RL pipeline.}
\label{fig:hapo_code}
\end{figure*}

\section{Theoretical Analysis of Entropy-Driven Modulation}

Below, we provide a formal theoretical analysis showing how entropy values drive each component and why HAPO systematically improves over DAPO by optimizing the gradient estimation process.

\subsection{Gradient Estimation under Entropy-Driven Modulation}
The ideal policy gradient is defined as:
$$\nabla_\theta J(\theta) = \mathbb{E}_{\tau \sim \pi_\theta}\bigg[\sum_t Q^\pi(s_t, a_t) \nabla_\theta \log \pi_\theta(a_t|s_t)\bigg]$$
In practice, the actual gradient estimator $\hat{g}$ differs from the ideal gradient. The mean squared error (MSE) of any gradient estimator admits the classic bias-variance decomposition:
$$\text{MSE}(\hat{g}) = \|\text{Bias}(\hat{g})\|^2 + \text{Var}(\hat{g})$$
where $\text{Bias}(\hat{g}) = \mathbb{E}[\hat{g}] - \nabla_\theta J(\theta)$. HAPO’s four components target these two terms from complementary angles: \textit{Adaptive Temperature Sampling} and \textit{Token-Level Group Average Advantage Estimation} primarily reduce $\|\text{Bias}\|^2$, while \textit{Differential Advantage Redistribution} and \textit{Asymmetric Adaptive Clipping} primarily reduce $\text{Var}(\hat{g})$.

\subsection{Adaptive Temperature Sampling: Reducing Bias}
Standard sampling uses $\pi_T(o_t|s_t) \propto \pi_\theta(o_t|s_t)^{1/T}$. HAPO sets $T_{i,t} = T_{\text{base}}(1+\tilde{h}_{i,t}\cdot\tau)$. In RLHF, the estimation quality depends on whether rollout samples cover critical reasoning paths. 

At a high-entropy position $t$, a critical action $a^*$ typically has low probability under $\pi_\theta$. In $G$ independent rollouts, the probability that $a^*$ is sampled at least once is $P(\text{sampled}) = 1 - (1 - \pi_T(a^*|s_t))^G$. When $T$ is fixed at a low value ($1/T > 1$), the exponentiation $\pi_\theta(a^*)^{1/T}$ suppresses already-small probabilities, making $P(\text{sampled})$ very low. This causes the advantage estimate to be biased toward dominant, potentially sub-optimal paths.

HAPO raises the temperature at high-entropy positions ($T_{i,t} > T_{\text{base}}$ when $\tilde{h}_{i,t} > 0$). Since $1/T_{i,t} < 1/T_{\text{base}}$, the suppression effect is reduced:
\begin{equation}
\begin{aligned}
\pi_{T_{i,t}}(a^*|s_t) &> \pi_{T_{\text{base}}}(a^*|s_t) \\
&\implies P^{\text{HAPO}} > P^{\text{DAPO}}
\end{aligned}
\end{equation}
More critical actions are covered, yielding a more representative advantage estimate and reducing coverage bias. Conversely, at low-entropy positions, lowering $T$ concentrates the distribution on the single viable action, reducing noise without introducing bias. HAPO breaks the global temperature trade-off:
$$T_{i,t}^{\text{high-ent}} > T_{\text{base}} > T_{i,t}^{\text{low-ent}}$$
thereby ensuring $\|\text{Bias}(\hat{g}^{\text{HAPO}})\|^2 < \|\text{Bias}(\hat{g}^{\text{DAPO}})\|^2$.

\subsection{Token-Level Group Average Advantage: Reducing Bias}
GRPO divides the loss by sequence length $|o_i|$, while DAPO divides by total tokens $N_{\text{tok}} = \sum_i |o_i|$. Both use sequence-level advantage $A_i = (R_i - \mu_R)/\sigma_R$. The gradient of DAPO's loss (simplified) is:
$$\hat{g}^{\text{DAPO}} = \frac{1}{N_{\text{tok}}} \sum_i A_i \cdot |o_i| \cdot \bar{\nabla}_i$$
where $\bar{\nabla}_i$ is the average per-token gradient. The weight of sequence $i$ is $A_i \cdot |o_i|$. Since negative samples are significantly longer ($|o_i^-| \gg |o_i^+|$), the balance is broken: $\sum_{i:A_i<0} |A_i| \cdot |o_i^-| \gg \sum_{i:A_i>0} |A_i| \cdot |o_i^+|$. 

This length-advantage coupling introduces a systematic negative bias. HAPO resolves this by normalizing across the global token pool:
$$A_{i,t}^{\text{tok}} = \frac{a_{i,t} - \mu_{\text{tok}}}{\sigma_{\text{tok}}}$$
By construction, $\sum_{(i,t)} A_{i,t}^{\text{tok}} = 0$ holds at the \textbf{token level}. Longer sequences contribute proportionally more tokens to both positive and negative sides, eliminating the bias while preserving the ability to learn from complex reasoning chains.

\subsection{Differential Advantage Redistribution: Reducing Variance}
Consider the per-token gradient $g_{i,t} = \hat{A}_{i,t} \cdot \nabla_\theta \log \pi_\theta(o_t|s_t)$. Its variance is $\text{Var}(g_{i,t}) = \hat{A}_{i,t}^2 \cdot F_t(\theta)$, where $F_t(\theta)$ is the Fisher Information. High-entropy positions have large $F_t$, yielding a low Signal-to-Noise Ratio (SNR).

HAPO redistributes advantage such that for high-entropy tokens ($\tilde{h} > 0$) with $r_{i,t} \notin [\gamma_L, \gamma_U]$, the advantage is amplified by $(1+\tilde{h})$. This boosts the SNR:
$$\text{SNR}_{i,t}^{\text{HAPO}} = \frac{(1+\tilde{h}_{i,t})^2 \cdot |\mathbb{E}[\nabla_\theta \log \pi_\theta]|^2}{F_t(\theta)}$$
The $(1+\tilde{h})^2$ factor compensates for the large $F_t$. Simultaneously, HAPO suppresses low-entropy tokens that have high SNR but low information. This reallocates the gradient budget to high-information tokens, reducing overall variance: $\text{Var}(\hat{g}^{\text{HAPO}}) < \text{Var}(\hat{g}^{\text{DAPO}})$.

\subsection{Asymmetric Adaptive Clipping: Reducing Variance}
The local KL divergence contribution from a token $o_t$ under a ratio change $\delta = |r_{i,t} - 1|$ is approximately $\Delta D_{\text{KL}}^{(t)}(\delta) \approx \delta^2 \cdot \pi_{\text{old}}(o_t)$. 

In DAPO, fixed clipping $[1-\epsilon_L, 1+\epsilon_R]$ causes high-entropy tokens ($\pi_{\text{old}}$ is small) to be clipped prematurely despite low KL cost, while low-entropy tokens ($\pi_{\text{old}}$ is large) undergo disproportionately large KL shifts. HAPO makes the clipping threshold a function of entropy:
$$\delta_R^{\max}(\tilde{h}) = \epsilon_R^{\text{base}}(1+\tilde{h}_{i,t}), \quad \delta_L^{\max}(\tilde{h}) = \epsilon_L^{\text{base}}(1-\tilde{h}_{i,t})$$
This ensures the effective trust region in KL space is approximately uniform across entropy levels:
$$\Delta D_{\text{KL}}^{\text{HAPO, high-ent}} \approx \Delta D_{\text{KL}}^{\text{HAPO, low-ent}}$$
This retains informative high-entropy gradients while properly constraining noisy low-entropy updates, further reducing variance.

\section{Practical Adoption}
\subsection{Hyperparameter Robustness}
HAPO uses a single configuration across all tested models (Qwen, LLaMA) and scales (1.5B--14B). Ablations in Appendix ~\ref{sec:additional_ablations} confirm that performance is stable across a wide range of values, requiring no per-model tuning.

\subsection{Implementation and Integration}
HAPO is easily integrated into existing RLHF pipelines (e.g., \texttt{verl}, \texttt{vLLM}) using simple element-wise operations. The core logic follows a unified principle: normalize entropy to $\tilde{h} \in [-1, 1]$ and scale base parameters accordingly. Implementation requires only minimal additions to the sampling, clipping, and advantage estimation modules.We provide full source code, configuration files, and a comprehensive setup guide at: \href{https://github.com/starriver030515/HAPO}{https://github.com/starriver030515/HAPO}.

\section{Example of Dual-Entropy Phenomenon}
\label{app:dual_entropy}
\begin{figure*}[!t]
    \centering
    \includegraphics[width=13.5cm]{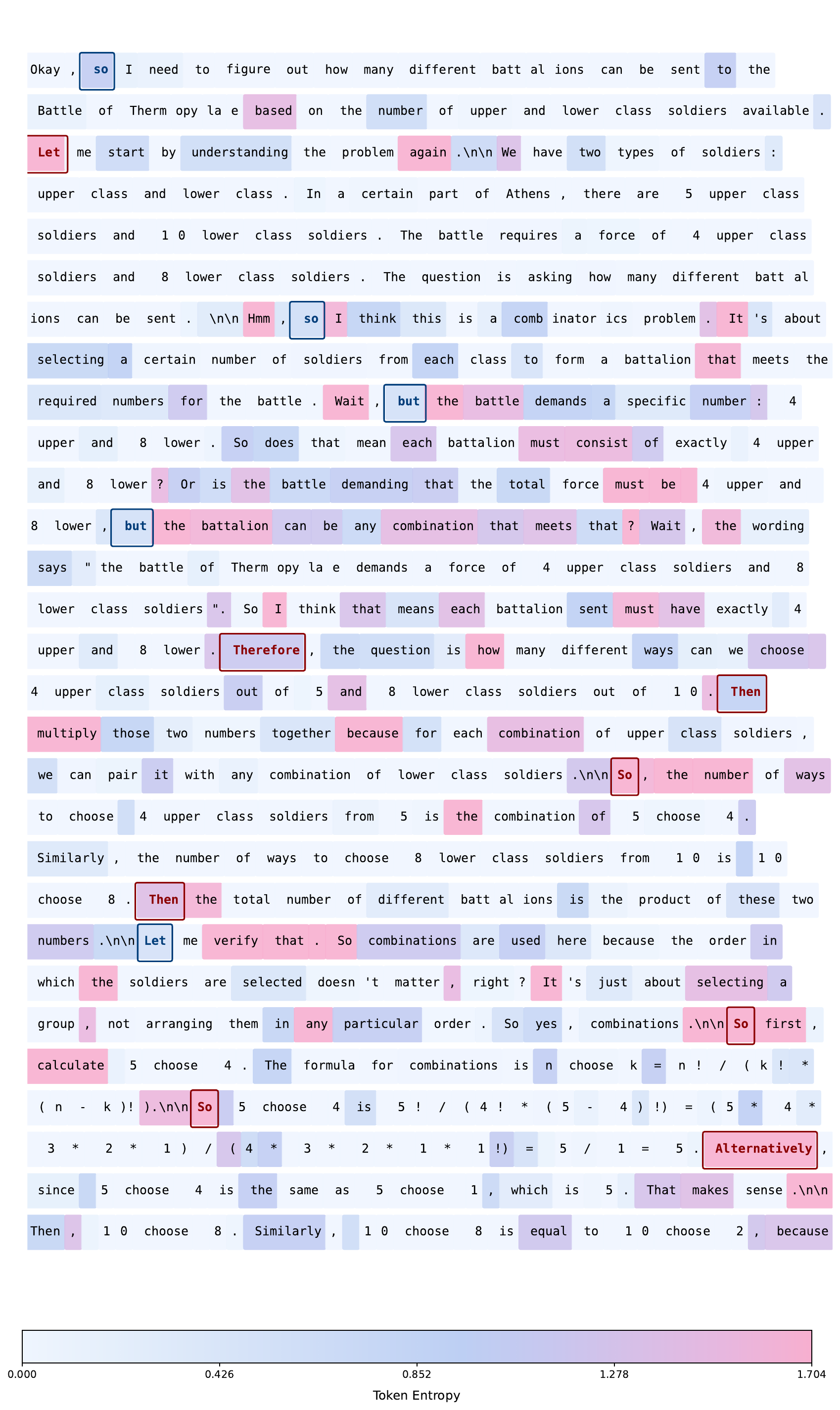}
    \caption{Dual-Entropy Example 1-part1}
    \label{fig:Dual_Entropy_1-1}
\end{figure*}

\begin{figure*}[!t]
    \centering
    \includegraphics[width=13.5cm]{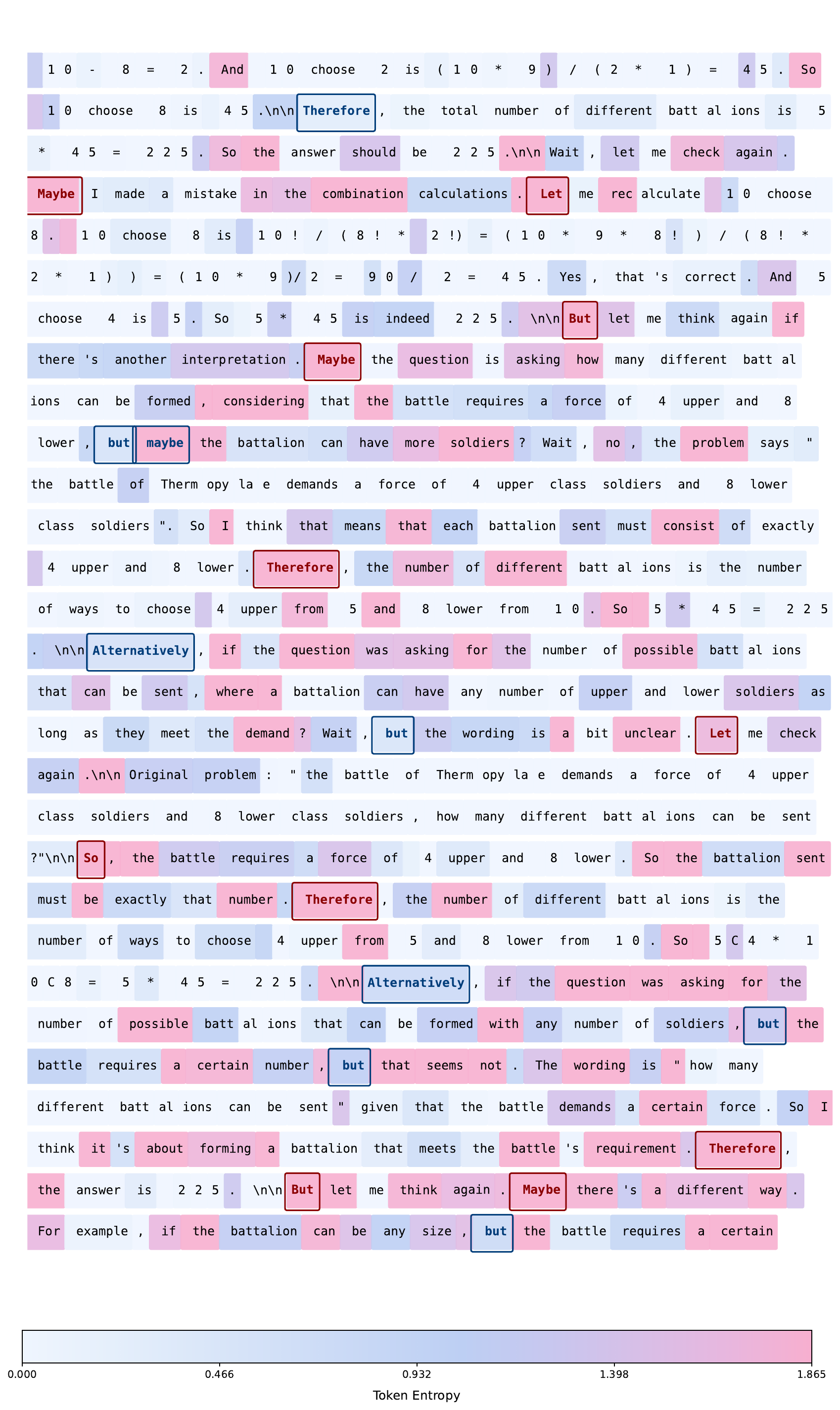}
    \caption{Dual-Entropy Example 1-part2}
    \label{fig:Dual_Entropy_1-2}
\end{figure*}

\begin{figure*}[!t]
    \centering
    \includegraphics[width=13.5cm]{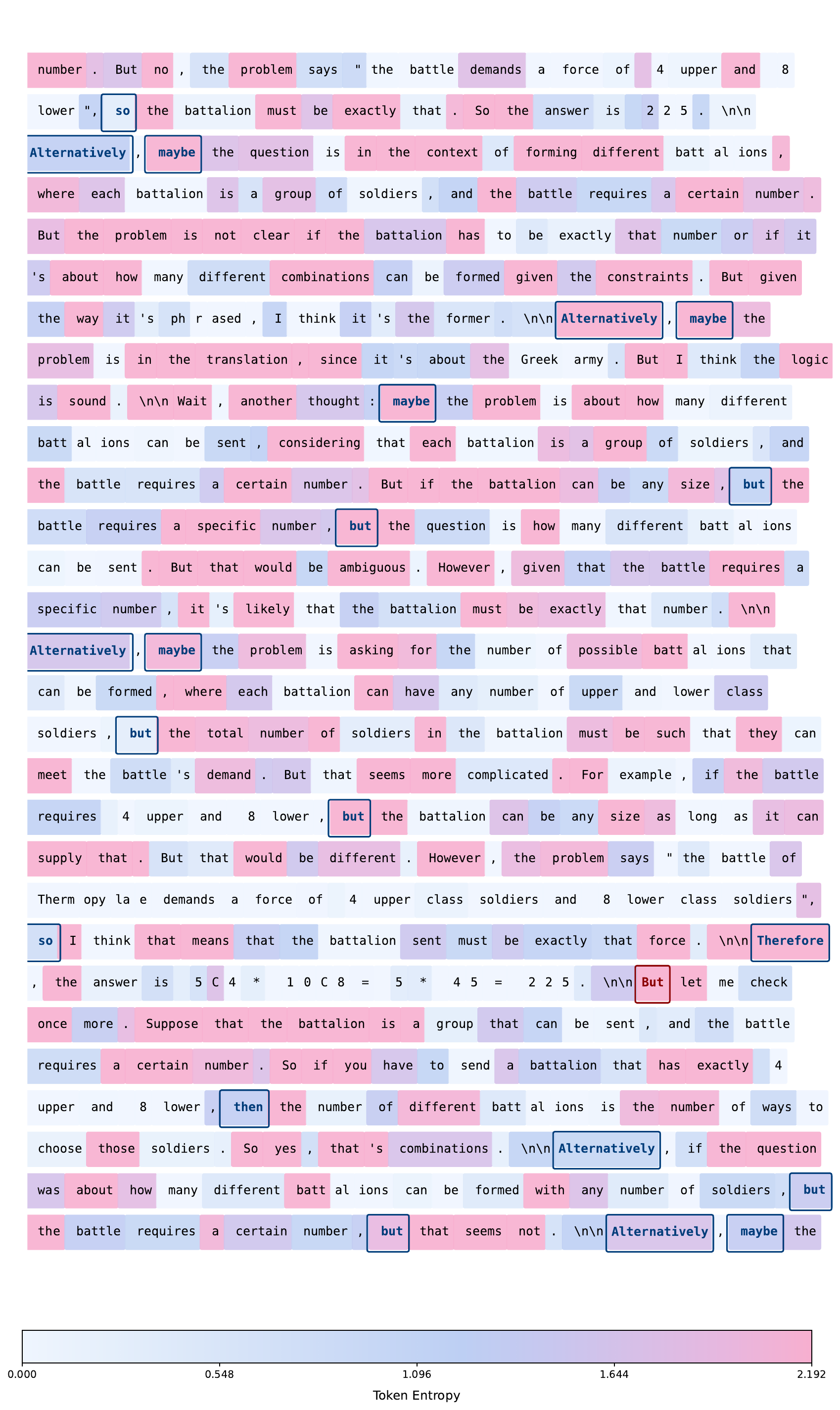}
    \caption{Dual-Entropy Example 1-part3}
    \label{fig:Dual_Entropy_1-3}
\end{figure*}

\begin{figure*}[!t]
    \centering
    \includegraphics[width=13.5cm]{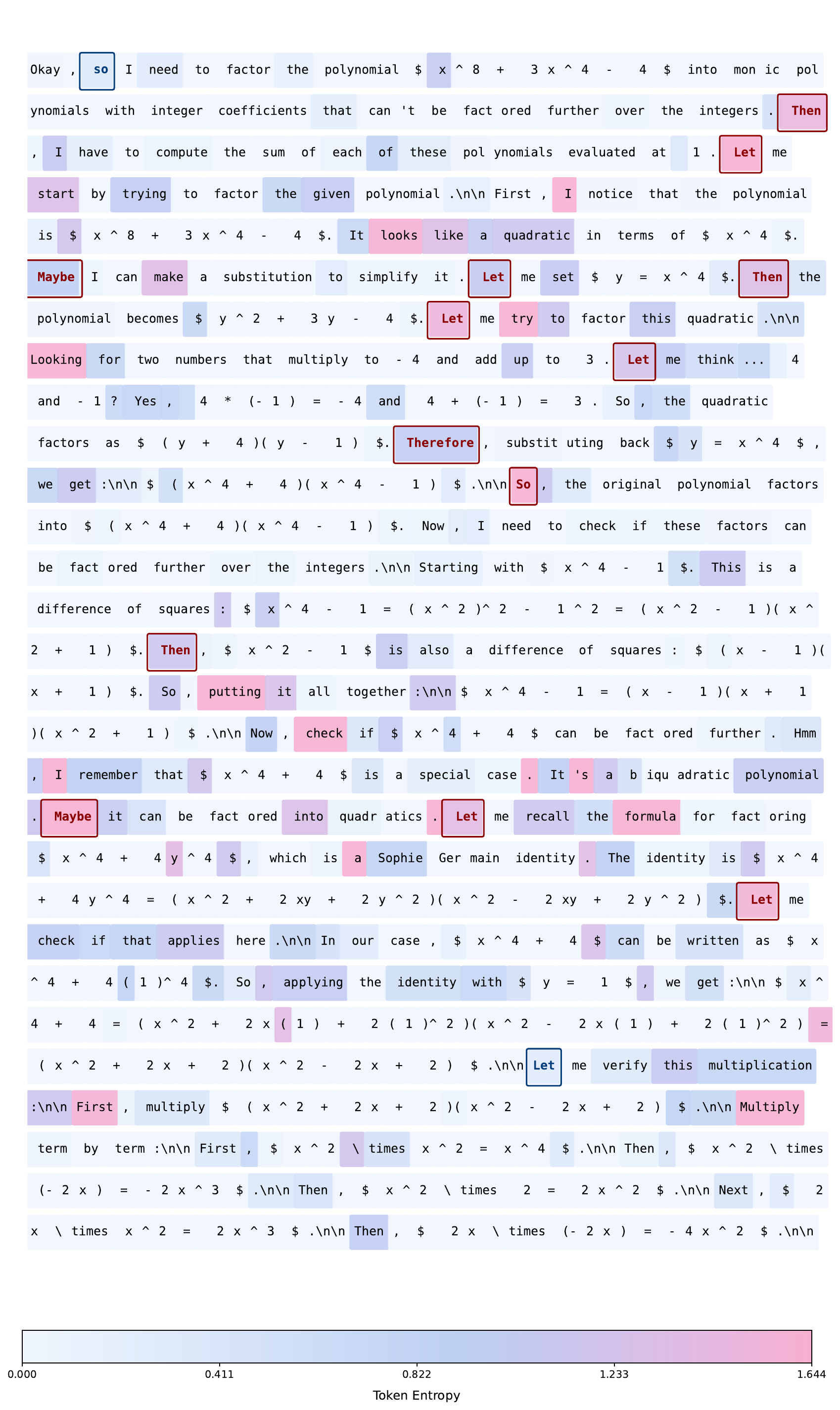}
    \caption{Dual-Entropy Example 2-part1}
    \label{fig:Dual_Entropy_2-1}
\end{figure*}

\begin{figure*}[!t]
    \centering
    \includegraphics[width=13.5cm]{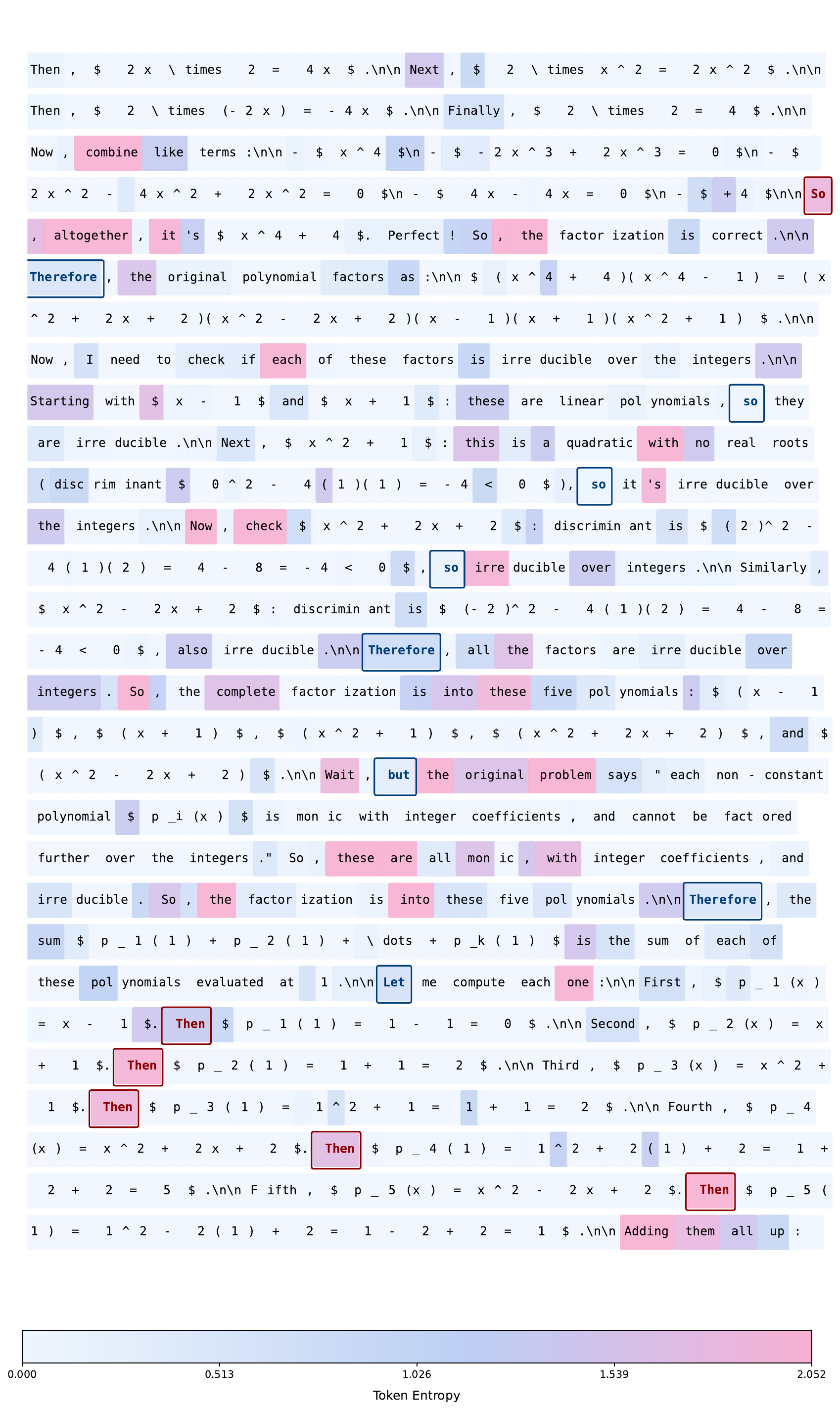}
    \caption{Dual-Entropy Example 2-part2}
    \label{fig:Dual_Entropy_2-2}
\end{figure*}

\begin{figure*}[!t]
    \centering
    \includegraphics[width=13.5cm]{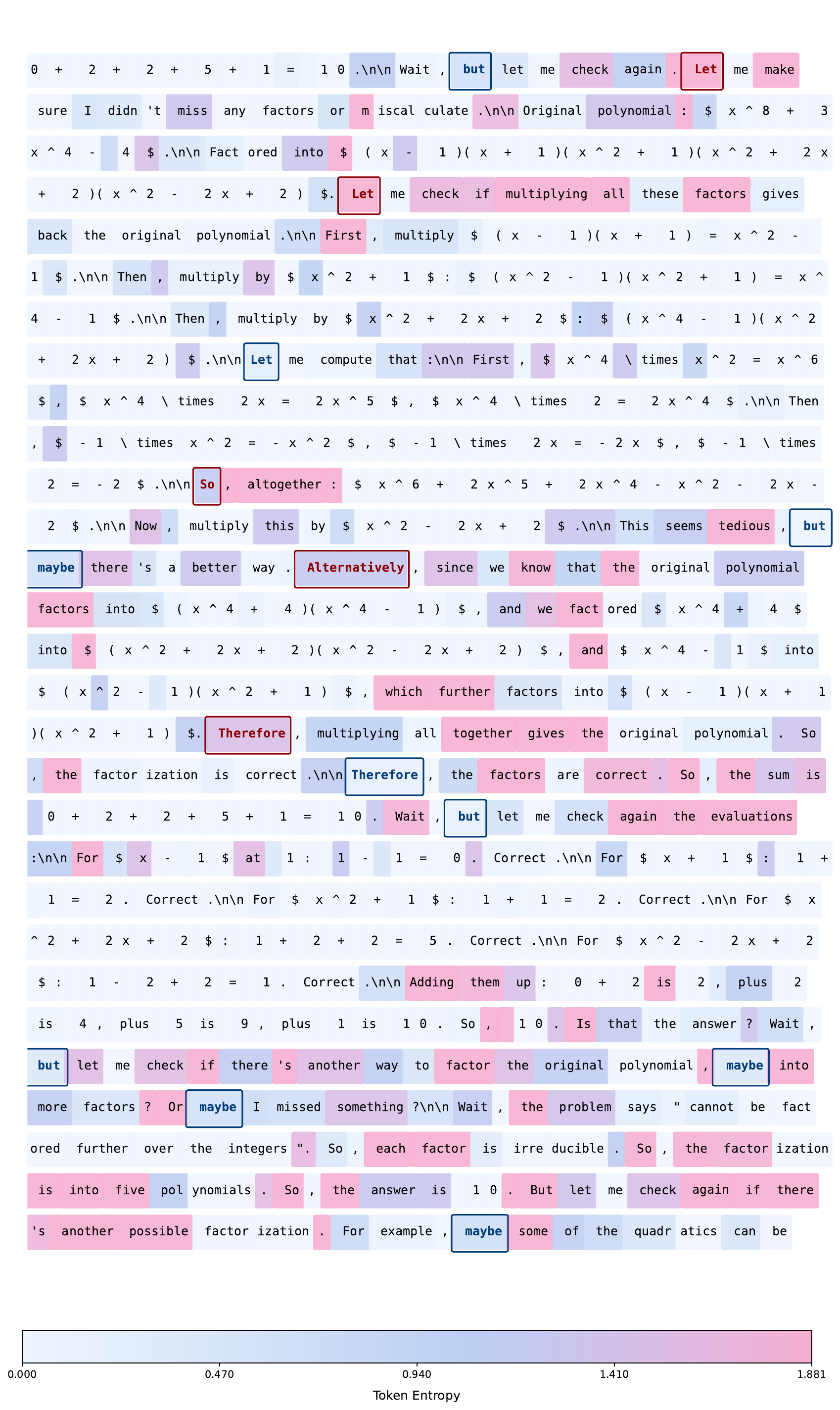}
    \caption{Dual-Entropy Example 2-part3}
    \label{fig:Dual_Entropy_2-3}
\end{figure*}

We show the 20 most frequent dual-entropy token pairs in Figure~\ref{fig:max_20_entropy}. These dual-entropy tokens are all critical tokens that determine reasoning paths, yet exhibit vastly different entropy values. We present concrete examples of the dual entropy phenomenon in Figure~\ref{fig:Dual_Entropy_1-1},~\ref{fig:Dual_Entropy_1-2},~\ref{fig:Dual_Entropy_1-3},~\ref{fig:Dual_Entropy_2-1},~\ref{fig:Dual_Entropy_2-2},~\ref{fig:Dual_Entropy_2-3}. We use blue and red to represent the low-entropy and high-entropy parts of dual-entropy, respectively. These visualizations demonstrate how tokens with identical semantic meanings but different surface forms exhibit drastically different entropy values during model generation. 

\end{document}